\documentclass[letterpaper]{article} % DO NOT CHANGE THIS
\usepackage{aaai2026}  % DO NOT CHANGE THIS
\usepackage{times}  % DO NOT CHANGE THIS
\usepackage{helvet}  % DO NOT CHANGE THIS
\usepackage{courier}  % DO NOT CHANGE THIS
\usepackage[hyphens]{url}  % DO NOT CHANGE THIS
\usepackage{graphicx} % DO NOT CHANGE THIS
\usepackage{natbib}  % DO NOT CHANGE THIS AND DO NOT ADD ANY OPTIONS TO IT
\usepackage{caption} % DO NOT CHANGE THIS AND DO NOT ADD ANY OPTIONS TO IT
\usepackage{algorithm}
\usepackage{algorithmic}

\usepackage{booktabs}
\usepackage{multirow}
\usepackage{makecell}
\usepackage{amsmath}
\usepackage{amssymb}
\usepackage{subcaption}
\usepackage{tcolorbox} 
\usepackage{newfloat}
\usepackage{listings}
\DeclareCaptionStyle{ruled}{labelfont=normalfont,labelsep=colon,strut=off} % DO NOT CHANGE THIS
\floatstyle{ruled}
\newfloat{listing}{tb}{lst}{}
\floatname{listing}{Listing}
\title{Detecting Emotional Dynamic Trajectories: An Evaluation Framework for Emotional Support in Language Models}
\author{
    Zhouxing Tan\textsuperscript{\rm 1}, Ruochong Xiong\textsuperscript{\rm 1}, Yulong Wan\textsuperscript{\rm 1}, Jinlong Ma\textsuperscript{\rm 2}, Hanlin Xue\textsuperscript{\rm 1}, Qichun Deng\textsuperscript{\rm 2}, Haifeng Jing\textsuperscript{\rm 1}, Zhengtong Zhang\textsuperscript{\rm 2}, Depei Liu\textsuperscript{\rm 1}, Shiyuan Luo\textsuperscript{\rm 2}, Junfei Liu\textsuperscript{\rm 1}\thanks{Corresponding author.}
}
\affiliations{
    \textsuperscript{\rm 1}National Engineering Research Center for Software Engineering, Peking University, Beijing, China\\
    \textsuperscript{\rm 2}Guangzhou Quwan Network Technology, Guangzhou, China\\
    \{tzhx, liujunfei\}@pku.edu.cn,\{2401110755, 2501110753, 2301210487\}@stu.pku.edu.cn, majinlong@52tt.com
}

\usepackage{bibentry}
\begin{document}

\maketitle
\begin{abstract}
Emotional support is a core capability in human-AI interaction, with applications including psychological counseling, role play, and companionship. However, existing evaluations of large language models (LLMs) often rely on short, static dialogues and fail to capture the dynamic and long-term nature of emotional support. To overcome this limitation, we shift from snapshot-based evaluation to trajectory-based assessment, adopting a user-centered perspective that evaluates models based on their ability to improve and stabilize user emotional states over time. Our framework constructs a large-scale benchmark consisting of 328 emotional contexts and 1,152 disturbance events, simulating realistic emotional shifts under evolving dialogue scenarios. To encourage psychologically grounded responses, we constrain model outputs using validated emotion regulation strategies such as situation selection and cognitive reappraisal. User emotional trajectories are modeled as a first-order Markov process, and we apply causally-adjusted emotion estimation to obtain unbiased emotional state tracking. Based on this framework, we introduce three trajectory-level metrics: Baseline Emotional Level (BEL), Emotional Trajectory Volatility (ETV), and Emotional Centroid Position (ECP). These metrics collectively capture user emotional dynamics over time and support comprehensive evaluation of long-term emotional support performance of LLMs. Extensive evaluations across a diverse set of LLMs reveal significant disparities in emotional support capabilities and provide actionable insights for model development.
\end{abstract}

\begin{links}
    \link{Code}{https://RuoChoXio.github.io/ETrajEval/}
\end{links}

\section{Introduction}

Emotional support conversation (ESC) \cite{liu2021towards,kang2024can} aims to alleviate emotional distress and offer constructive guidance through empathetic dialogue. With advances in large language models (LLMs), ESC has expanded beyond emotion recognition and generation to include broader human-centric tasks such as role-playing \cite{shanahan2023role}, psychological companionship\cite{sorin2024large}, and casual chatting\cite{llanes2024developing}. Effective ESC not only reduces negative emotions but also helps sustain positive emotional states through consistent, high-quality interaction. Ideally, such support should unfold through long-term \cite{zhong2024memorybank} and dynamic \cite{castillo2024beyond} conversations, where long-term refers to sustained multi-turn dialogue that enables the model to accumulate dialogue context and understand the user’s evolving state, and dynamic refers to open-ended exchanges shaped by both the user and the model, with flexibility to adapt to external emotional influences.

While existing evaluation methods have advanced our understanding of LLMs’ emotional capabilities, these approaches exhibit two major limitations: 
\textbf{(a). Overemphasis on model-centric response quality}. Most evaluations rely on benchmark datasets and assess generation quality from the model’s perspective, which introduces subjectivity . (a1). In varied dialogue contexts, single-reference answers are insufficient to represent the diversity of valid replies. (a2). Emotional support is inherently ambiguous and open-ended, and reference responses in datasets may not reflect optimal or unbiased support, limiting the fairness and robustness of evaluation. \cite{smith2022human,wang2023large} \textbf{(b). Lack of long-term and dynamic interaction}. Current evaluations are typically limited to short, static conversations in controlled settings, failing to consider evolving user emotional shifts. This overlooks essential aspects of real emotional support, including multi-turn continuity and contextual adaptation \cite{madani2024steering}. Without trajectory-level metrics, it is difficult to assess whether the model truly improves or maintains the user’s emotional well-being over time.

To better evaluate the emotional support capabilities of LLMs, we adopt a \textbf{user-centered perspective} that focuses on the user’s emotional trajectory throughout interaction. A model is considered emotionally competent if it can consistently improve and stabilize the user’s emotional state over time. Grounded in psychological theory \cite{gross2015emotion,aldao2010emotion,hocker2018interpersonal}, our framework addresses key limitations as follows.

To address Problem (a1), we construct a benchmark comprising 328 emotional contexts and 1,152 disturbance events to simulate realistic emotional shifts and evaluate model adaptability under evolving scenarios. To address Problem (a2), we constrain model responses using emotion regulation strategies grounded in psychological theory, such as situation selection and cognitive reappraisal, to encourage supportive behaviors aligned with validated therapeutic principles. To address Problem (b), we simulate long-term dynamic interactions involving repeated emotional disturbances. User emotional trajectories are modeled as a first-order Markov process, and causally-adjusted emotion estimation is applied to enable unbiased tracking of emotional states. Based on this framework, we propose three trajectory-level metrics: Baseline Emotional Level (BEL), Emotional Trajectory Volatility (ETV), and Emotional Centroid Position (ECP). These metrics jointly characterize dynamics of user emotional states and serve as indicators for evaluating emotional fluctuations and stability over time.

\begin{figure*}[ht]
\centering
\includegraphics[width=0.99\textwidth]{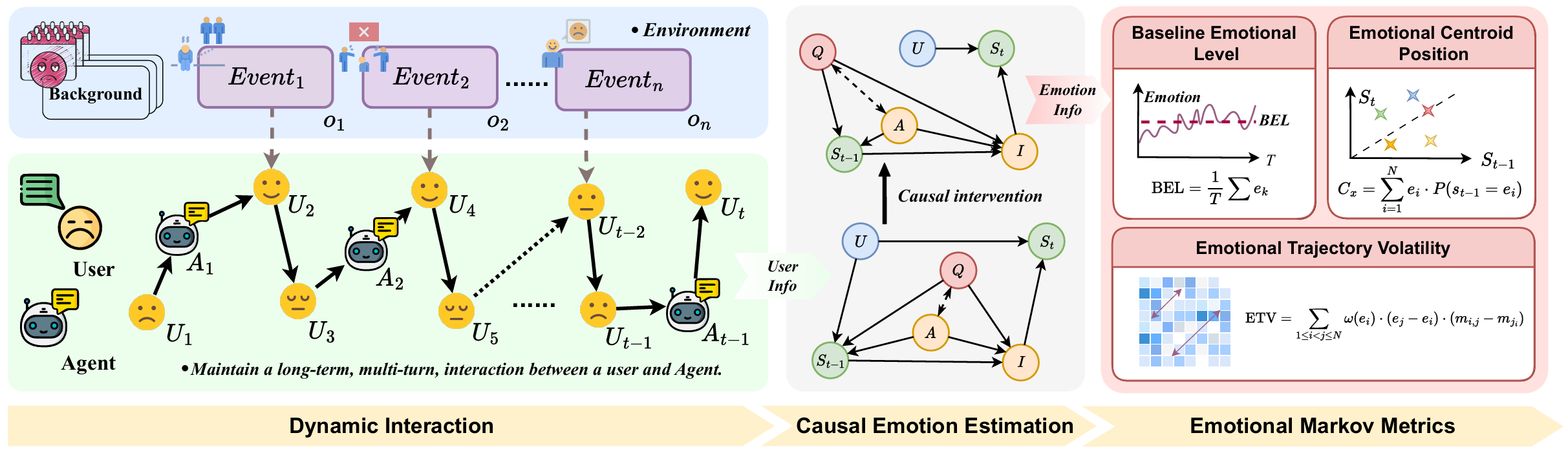}
\caption{Overview of our evaluation framework for emotional support in long-term dialogues. It includes three modules: dynamic user-agent interaction under emotional events, causal emotion estimation based on Markov modeling, and three trajectory-level metrics including Baseline Emotional Level (BEL), Emotional Trajectory Volatility (ETV), and Emotional Centroid Position (ECP).}
\label{framework}
\end{figure*}

Together, these components form a dynamic evaluation framework comprising three pillars: Evaluation Environment, Dynamic Interaction, and Trajectory-Based Metrics (Fig.~\ref{framework}). Our main contributions are as follows:
\begin{itemize}
    \item \textbf{A user-centered evaluation framework}: We propose a dynamic, long-term evaluation framework that tracks user emotional trajectories using a Markov process and causally-adjusted estimation. It introduces three trajectory-level metrics (BEL, ETV, ECP) and is supported by formal theoretical justification.
    \item \textbf{A realistic benchmark with psychologically grounded design}: We build a large-scale benchmark of 328 emotional contexts and 1,152 disturbance events, and constrain model responses using validated emotion regulation strategies from psychology.
    \item \textbf{Empirical Insights}: Through extensive evaluations across diverse LLMs, we uncover significant disparities in their long-term emotional support capabilities and offer actionable insights for developing more emotionally supportive LLMs.
\end{itemize}

\section{Related Work}

The evaluation of a model's emotional support capabilities can be categorized into automated and manual approaches. Early automated methods focused on the quality of generated text, using traditional metrics like BLEU and ROUGE to compare model outputs with reference answers \cite{rashkin2018towards}. However, these metrics show a weak correlation with human perceptions of empathy, creativity, and overall dialogue quality. Subsequent research shifted to assessing generation quality through metrics such as fluency, coherence, naturalness, empathy and so on \cite{xu2024multi, tu2024charactereval, zhang2024feel}. Such evaluations typically rely on single-turn Q\&A \cite{afzoon2024persobench, zhao2024esc, chen2024socialbench} or short, multi-turn dialogues (fewer than 10 turns) \cite{yuan2025dmt, tamoyan2024llm, feng2025emocharacter} to generate responses, which are then directly scored using an LLM-as-a-judge framework \cite{tu2024charactereval, zhou2025characterbench}. This direct scoring method, however, lacks transparency and may introduce subjective bias. An alternative automated approach indirectly assesses emotional support by evaluating the model's emotional intelligence \cite{sabour2024emobench, paech2023eq, chen2024emotionqueen}. This involves deconstructing the assessment into separate tests of the model's capabilities in emotion recognition, understanding, and generation. While this method effectively gauges foundational abilities, it struggles to measure holistic performance in the context of multi-turn conversations.

Manual evaluation, on the other hand, involves hiring experts or crowd-workers to assess and annotate model responses through techniques like pairwise comparisons \cite{wu2025raiden}, ranking \cite{rashkin2018towards}, and scoring on multi-dimensional scales \cite{chen2023soulchat,welivita2024large}. These methods demonstrate alignment with human intent but are costly, time-consuming, and still subject to a degree of subjectivity. Moreover, a significant limitation shared by both automated and manual evaluations is their narrow focus on the quality of model's responses. And they often overlook the dynamic and contextual relationships that develop over the course of a multi-turn dialogue.

\section{Affective Evaluation Environment and Interaction Design}
To evaluate LLMs’ emotional support capabilities in realistic settings, we design a dynamic environment simulating users’ emotional trajectories under distress. It includes user scenarios and interaction patterns grounded in psychological theory, featuring (1) emotionally charged user contexts prone to downturns, and (2) model-side constraints based on emotion regulation strategies. We further introduce disturbance events to stress-test support consistency and formalize the interaction as a state-based trajectory framework, enabling structured evaluation of long-term affective support.

\subsection{User-Side: Emotional Distress Scenarios}
We categorize emotional distress into four domains: professional and social roles, intimate relationships, personal struggles, and life circumstances, drawing from psychological and conflict theory literature \cite{hocker2018interpersonal,kassin2024social}. Within these domains, we define 14 scenarios that capture common triggers of emotional suppression, conflict, or disruption. These scenarios form the foundation for constructing dynamic emotional trajectories in our evaluation. Detailed descriptions are provided in the extended version.

\subsection{Model-Side: Psychological Support Constraints}
To guide model behavior in emotionally supportive interactions, we adopt emotion regulation strategies grounded in psychological theory. Our framework draws on Gross’s process model of regulation \cite{gross2015emotion}, which defines sequential stages; Bonanno and Burton’s concept of regulatory flexibility \cite{bonanno2013regulatory}, which emphasizes adaptive strategy use; and Aldao et al.’s meta-analysis \cite{aldao2010emotion}, which highlights the varied effectiveness of strategies across contexts. These theories jointly provide a foundation for assessing the emotional intelligence, adaptability, and strategic competence of language models.

We operationalize this framework through six evaluation dimensions: \textbf{Situation Selection (SitSel)}, which involves detecting emotional triggers and guiding users to supportive environments; \textbf{Situation Modification (SitMod)}, which focuses on suggesting changes to external or interpersonal conditions; \textbf{Attentional Deployment (AttDep)}, which aims to redirect attention to mitigate rumination; \textbf{Cognitive Change (CogChg)}, which refers to identifying cognitive distortions and offering constructive reappraisals; \textbf{Response Modulation (ResMod)}, which entails recommending actions to manage emotional responses; and \textbf{Emotion Regulation Flexibility (ERFlex)}, which emphasizes adapting strategies based on user state and context.

All data undergo rigorous human inspection, with distributional statistics reported in the extended version.

\subsection{Perturbations in Emotional Interaction}
To assess the model’s ability to sustain effective emotion regulation over extended interactions, we introduce emotion aggravation events which simulated negative surges that disrupt otherwise stabilized dialogues. We inject carefully designed triggers to emulate real-world emotional escalation. These events align with the narrative and psychological trajectory of the scenario. For instance, in a context involving regret over past life choices, a realistic aggravation might involve comparisons to more “successful” peers at a family gathering. Such events enhance ecological validity, increase evaluation difficulty, and enable robustness testing. They are randomly interleaved into dialogues to construct adversarial multi-turn sequences that challenge the model’s long-term emotional support capabilities.

\subsection{Formalization of the Dynamic Interaction Process}
For the environment $B$ defined in the previous section, we construct a dynamic multi-turn interaction process involving a user $Q$ and an emotion support model $A$. At each turn $i$, the dialogue history is denoted by: $H_i = (q_1, a_1, q_2, a_2, ..., q_T, a_T)$, where $q_i$ and $a_i$ are the user’s utterance and the agent’s response at turn $i$, respectively. All interactions unfold within the shared environment $B$.

To simulate real-world disruptions, we introduce a set of disturbance events $O = \{o_1, ..., o_k\}$ aligned with the background logic of $B$. Each event $o_m \in O$ includes its content and a predefined trigger point. After the agent produces response $a_i$, an event $o$ may be triggered and presented to the user, modifying their informational state.

As a result, the user’s next utterance $q_{i+1}$ is influenced by both the dialogue history $H_i$ and the triggered events, formalized as: $q_{i+1, O_i} = \text{User}(H_i, B, O_i)$, where $O_i \subseteq O$ is the set of events observed up to turn $i$. In the following turn, the agent receives $q_{i+1, O_i}$ without direct access to $o$, and must infer the updated context from the user’s behavior. The agent’s response is then given by: $a_{i+1} = A(q_{i+1, O_i} \mid H_i, B)$.

\section{User Perspective Dynamic Trajectories Metrics}

\subsection{Emotional Markov Metric}
Due to environmental variability and diverse model-user interactions, defining a unified standard to evaluate the quality of a model's emotional support responses is challenging. To address this, we propose metrics that evaluate the model's emotional support capability by quantifying the dynamic evolution of user emotional states over long-term interactions, grounded in the principle that effective support should guide users toward or sustain positive moods.

We model emotion as a continuous spectrum $\mathcal{E} \subset \mathbb{R}$ ranging from highly negative to highly positive, and discretize it into $N$ ordered, disjoint intervals $\{\mathcal{E}_1, \mathcal{E}_2, \dots, \mathcal{E}_N\}$. This yields a finite emotional state space $\mathcal{S} = \{e_1, e_2, \dots, e_N\}$, where smaller $s \in \mathcal{S}$ indicates more negative emotion and larger $s$ indicates more positive. Let $s_t$ denote the user’s emotional state after the $t$-th interaction. Assuming a first-order Markov process, we model the emotional trajectory as: $P(s_t \mid s_{t-1}, \dots, s_0) = P(s_t \mid s_{t-1})$. This defines a Markov chain $(s_0, s_1, \dots, s_T)$. Based on this, we introduce three complementary metrics to capture emotional level, transition asymmetry, and centroid shift.

\subsubsection{Baseline Emotional Level (BEL)}  
BEL quantifies the average emotional level of the user after a $T$-turn interaction with the model. Given an observed emotion sequence $(s_1, \dots, s_T)$, BEL is defined as:
\begin{equation}
BEL = \frac{1}{T} \sum_{t=1}^{T} s_t.
\end{equation}
A higher BEL indicates that the model helps maintain a more positive average emotional state.

While BEL, as a mean-level metric, fails to reveal the internal dynamics of emotional change, we introduce the emotional transition matrix to construct new metrics that better reflect the efficiency and magnitude of user emotional improvement. The emotional transition matrix $M$ is an $N \times N$ matrix where each entry $m_{i,j}$ is defined as the conditional probability of the user's emotional state transitioning from $i$ to $j$ after a single interaction with the model: $m_{i,j} = P(s_t=e_j | s_{t-1}=e_i), \quad \forall e_i,e_j \in \mathcal{E}$. Specifically, for each predefined evaluation environment, we generate a complete dialogue of $T$ interaction turns, which yields an observed emotional state sequence $(s_0, s_1, \dots, s_T)$, where $s_t \in \mathcal{E}$. Based on this sequence, we estimate the emotional transition matrix $M$ corresponding to that context. The extended version provides the detailed methodology for constructing the transition matrix and a justification for its validity as a metric. Building on this, we propose the following two new evaluation metrics.

\subsubsection{Emotional Trajectory Volatility (ETV)}  
This metric quantifies the model’s ability to elevate users from negative emotional states and prevent regressions from positive ones. It focuses on the asymmetry between upward ($e_i \to e_j,\, j > i$) and downward ($e_j \to e_i$) transitions. We define the \textit{Asymmetric Transition Advantage} as the net transition gain: $d_{i,j} = m_{i,j} - m_{j,i}, \quad \text{for } e_j > e_i$. To account for transition difficulty, we introduce a \textit{Transition Value Weight} as the state distance $(e_j - e_i)$. Additionally, we define a \textit{State Importance Weight} $\omega(e_i)$ to capture the higher utility of improving from lower emotional states, based on diminishing returns. Combining these elements, we define the ETV as:
\begin{equation}
ETV = \sum_{1 \le i < j \le N} \omega(e_i) \cdot (e_j - e_i) \cdot (m_{i,j} - m_{j,i}).
\end{equation}
A higher ETV indicates stronger upward regulation and resilience against emotional decline.

\subsubsection{Emotional Centroid Position (ECP)}  

Emotional Centroid Position (ECP): To better visualize the magnitude of emotional transitions in multi-turn dialogues, we design the Emotional Centroid Position (ECP), a metric based on the transition matrix $M$. We define the ECP as the expected position of a state transition under the probabilistic model formed by the dialogue's empirical initial state distribution and the transition matrix $M$. In the extended version, we demonstrate that the two coordinates of this centroid, $C_{\mathcal{E}}=(C_x,C_y)$, have a clear geometric interpretation: the abscissa $C_x$ represents the average initial emotional value \textit{before} transitions occur, while the ordinate $C_y$ represents the average emotional value achieved \textit{after} the transitions. The specific computation is as follows:
\begin{small}

\begin{equation}
\begin{gathered}
C_x = \mathbb{E}[s_{t-1}] = \sum_{i=1}^{N} e_i \cdot P(s_{t-1} = e_i), \\
C_y = \mathbb{E}[s_t] = \sum_{i=1}^{N} \sum_{j=1}^{N} e_j \cdot m_{ij} \cdot P(s_{t-1} = e_i).
\end{gathered}
\end{equation}
\end{small}
In this formulation, $P(s_{t-1} = e_i)$ denotes the empirical probability that a transition begins in state $e_i$ within the given dialogue. 

\begin{table*}[t]
\small
\setlength{\tabcolsep}{4pt}
\renewcommand{\arraystretch}{0.95}
\centering
\begin{tabular}{l|cc|cc|cc|cc|cc|cc|cc}
\toprule
\multirow{2}{*}{\textbf{LLMs}} 
& \multicolumn{2}{c|}{\textbf{Overall}}
& \multicolumn{2}{c|}{\textbf{CogChg}} 
& \multicolumn{2}{c|}{\textbf{SitMod}} 
& \multicolumn{2}{c|}{\textbf{AttDep}} 
& \multicolumn{2}{c|}{\textbf{ERFlex}} 
& \multicolumn{2}{c|}{\textbf{SitSel}} 
& \multicolumn{2}{c}{\textbf{ResMod}} 
\\
\cmidrule(lr){2-3} \cmidrule(lr){4-5} \cmidrule(lr){6-7} \cmidrule(lr){8-9} 
\cmidrule(lr){10-11} \cmidrule(lr){12-13} \cmidrule(lr){14-15}
& EN & ZH & EN & ZH & EN & ZH & EN & ZH & EN & ZH & EN & ZH & EN & ZH \\
\midrule
\multicolumn{15}{l}{\textit{Closed-sourced LLMs}} \\
\midrule
O3-2025-04-16 & 43.98 & 39.22 & 48.64 & 40.31 & 38.54 & 36.09 & \underline{50.81} & 44.94 & 37.87 & 34.35 & 43.27 & 39.67 & 45.31 & 41.29 \\
Gemini-2.5-Pro & 42.65 & 37.41 & 44.18 & 33.91 & 43.65 & 37.51 & 37.12 & 39.48 & 38.47 & 30.46 & 42.18 & 40.65 & 49.00 & 44.80 \\
Claude-Opus-4 & 41.43 & 34.68 & 46.22 & 34.16 & 40.12 & 29.09 & 41.99 & 35.26 & 31.14 & 31.97 & 46.27 & 37.22 & 41.47 & 43.15 \\
Doubao-Seed-1.6 & 47.22 & 42.36 & 47.72 & 37.87 & 46.13 & 40.98 & 46.49 & 43.71 & 42.96 & 38.65 & 50.27 & \underline{47.88} & \underline{50.17} & 48.34 \\
Doubao-1.5-Pro-Character & 34.90 & 37.51 & 33.68 & 35.78 & 31.42 & 34.11 & 37.62 & 42.73 & 36.54 & 35.90 & 37.56 & 41.34 & 34.68 & 37.94 \\
ChatGPT-4o-Latest & \underline{48.86} & 43.84 & \textbf{52.45} & 41.02 & 44.51 & \textbf{41.78} & \textbf{53.41} & 48.48 & 44.58 & 39.27 & \underline{51.21} & \textbf{48.61} & 47.44 & 46.73 \\
GLM-4-Plus & 42.92 & 36.39 & 44.44 & 36.94 & 39.32 & 32.61 & 45.66 & 39.34 & 39.81 & 34.99 & 44.13 & 36.53 & 45.20 & 39.49 \\
ChatGLM-4 & 40.14 & 33.83 & 40.29 & 31.87 & 37.33 & 32.15 & 43.15 & 36.93 & 37.75 & 31.90 & 41.22 & 34.26 & 42.52 & 37.92 \\
Grok-4 & 39.73 & 31.55 & 42.41 & 28.60 & 37.19 & 29.31 & 43.73 & 36.15 & 35.68 & 26.82 & 37.56 & 35.63 & 41.92 & 35.82 \\
\midrule
\multicolumn{15}{l}{\textit{Open-sourced LLMs}} \\
\midrule
Phi-4-14B & 44.69 & 37.42 & 44.10 & 33.58 & 42.41 & 32.56 & 47.48 & 42.13 & 42.15 & 37.08 & 45.71 & 42.62 & 47.87 & 40.84 \\
DeepSeek-V3 & 44.69 & 35.43 & 45.30 & 34.88 & 40.27 & 28.71 & 46.54 & 42.53 & 45.12 & 33.25 & 46.32 & 40.61 & 46.19 & 36.17 \\
DeepSeek-R1 & 47.57 & \textbf{45.32} & 48.87 & \underline{43.43} & 43.57 & 40.77 & 46.22 & \textbf{50.64} & \underline{46.55} & \textbf{47.01} & \textbf{52.90} & 47.88 & 48.23 & 45.31 \\
Qwen3-235B-A22B & 47.18 & 40.40 & 47.78 & 38.08 & \underline{46.52} & 36.57 & 49.50 & 42.33 & 45.35 & 37.12 & 48.20 & 42.72 & 45.94 & \underline{48.67} \\
Mistral-3.2-24B-Instruct & 33.93 & 27.93 & 32.71 & 27.63 & 30.26 & 24.09 & 36.13 & 29.46 & 31.06 & 26.25 & 36.98 & 30.15 & 38.90 & 31.93 \\
Kimi-K2-Preview & \textbf{49.00} & \underline{45.11} & \underline{49.96} & \textbf{47.57} & \textbf{49.34} & \underline{41.74} & 48.55 & \underline{50.00} & 43.86 & \underline{41.04} & 49.81 & 45.73 & \textbf{52.11} & 45.19 \\
Qwen3-8B & 46.56 & 38.85 & 47.97 & 37.02 & 45.67 & 33.74 & 45.79 & 46.75 & 44.65 & 33.41 & 47.52 & 42.03 & 47.46 & 43.98 \\
Qwen3-32B & 46.87 & 40.88 & 47.32 & 38.03 & 44.17 & 35.98 & 47.81 & 46.34 & \textbf{46.69} & 36.36 & 47.37 & 42.73 & 48.79 & \textbf{49.99} \\
Llama-3.1-70B-Instruct & 42.94 & 31.07 & 42.15 & 30.23 & 41.12 & 30.13 & 46.55 & 30.47 & 43.69 & 30.38 & 44.51 & 33.40 & 40.98 & 32.66 \\
\bottomrule
\end{tabular}
\caption{Comparison of Baseline Emotional Level (BEL) across different models. For each case, the model with the highest BEL is marked in \textbf{bold}, and the second-highest is \underline{underlined}, indicating their relative effectiveness in maintaining users’ emotional baseline.}
\label{BEL}
\end{table*}

\begin{table*}[ht]

\small
\setlength{\tabcolsep}{4pt}
\renewcommand{\arraystretch}{0.95}
\centering
\begin{tabular}{l|cc|cc|cc|cc|cc|cc|cc}
\toprule
\multirow{2}{*}{\textbf{LLMs}} 
& \multicolumn{2}{c|}{\textbf{Overall}}
& \multicolumn{2}{c|}{\textbf{CogChg}} 
& \multicolumn{2}{c|}{\textbf{SitMod}} 
& \multicolumn{2}{c|}{\textbf{AttDep}} 
& \multicolumn{2}{c|}{\textbf{ERFlex}} 
& \multicolumn{2}{c|}{\textbf{SitSel}} 
& \multicolumn{2}{c}{\textbf{ResMod}} 
\\
\cmidrule(lr){2-3} \cmidrule(lr){4-5} \cmidrule(lr){6-7} \cmidrule(lr){8-9} 
\cmidrule(lr){10-11} \cmidrule(lr){12-13} \cmidrule(lr){14-15}
& EN & ZH & EN & ZH & EN & ZH & EN & ZH & EN & ZH & EN & ZH & EN & ZH \\
\midrule
\multicolumn{15}{l}{\textit{Closed-sourced LLMs}} \\
\midrule
O3-2025-04-16 & 20.17 & 14.89 & 19.91 & \textbf{14.74} & 18.23 & 12.45 & 28.09 & 20.39 & 15.26 & 11.54 & 18.55 & 13.87 & 22.65 & 17.95 \\
Gemini-2.5-Pro & 18.95 & 13.39 & 18.65 & 9.47 & 19.80 & \underline{14.43} & 21.01 & 19.05 & 15.57 & 10.45 & 17.99 & 13.26 & 20.84 & 15.85 \\
Claude-Opus-4 & 19.89 & 10.81 & 20.15 & 9.64 & 20.13 & 7.47 & 26.10 & 14.53 & 13.12 & 7.56 & \underline{21.03} & 13.26 & 19.38 & 14.83 \\
Doubao-Seed-1.6 & 21.55 & 15.31 & 19.92 & 11.68 & 21.97 & 13.81 & 25.38 & 19.25 & 18.46 & 14.74 & \textbf{21.72} & \underline{15.69} & 22.94 & \underline{19.48} \\
Doubao-1.5-Pro-Character & 17.72 & 14.48 & 15.62 & 11.66 & 16.13 & 12.53 & 23.45 & 21.40 & 16.46 & 15.02 & 17.71 & 12.08 & 19.11 & 16.88 \\
ChatGPT-4o-Latest & \underline{21.99} & \underline{15.40} & \textbf{22.09} & 13.26 & 20.75 & 13.32 & \textbf{29.02} & 20.74 & 19.23 & 13.55 & 20.62 & 15.37 & 21.30 & 18.56 \\
GLM-4-Plus & 21.00 & 12.56 & 20.55 & 12.63 & 18.22 & 9.88 & 25.29 & 17.72 & 20.03 & 11.63 & 19.92 & 9.77 & \underline{23.68} & 15.13 \\
ChatGLM-4 & 20.25 & 11.56 & 19.58 & 8.90 & 19.15 & 10.29 & 25.05 & 15.14 & 17.53 & 11.57 & 19.25 & 9.32 & 22.17 & 16.25 \\
Grok-4 & 18.85 & 8.73 & 17.83 & 6.71 & 19.19 & 7.30 & 25.40 & 11.45 & 15.53 & 5.61 & 15.61 & 9.79 & 20.51 & 13.51 \\
\midrule
\multicolumn{15}{l}{\textit{Open-sourced LLMs}} \\
\midrule
Phi-4-14B & 21.37 & 13.21 & 21.27 & 9.94 & 19.66 & 11.24 & 25.65 & 19.12 & 18.99 & 13.89 & 20.12 & 13.60 & 23.64 & 14.33 \\
DeepSeek-V3 & 20.32 & 12.02 & 19.52 & 10.77 & 18.82 & 7.57 & 25.04 & 17.32 & \underline{20.39} & 10.77 & 18.98 & 14.22 & 20.49 & 14.32 \\
DeepSeek-R1 & 20.38 & 15.30 & 18.27 & 13.20 & 20.68 & 14.26 & 23.76 & 21.70 & 19.77 & \underline{15.37} & 20.03 & 13.25 & 20.97 & 15.92 \\
Qwen3-235B-A22B & 21.13 & 14.02 & 19.40 & 11.89 & \underline{23.10} & 13.66 & 26.46 & 19.38 & 19.99 & 11.88 & 19.86 & 11.67 & 18.47 & 17.35 \\
Mistral-3.2-24B-Instruct & 15.93 & 7.01 & 13.41 & 6.64 & 14.88 & 4.19 & 23.75 & 9.58 & 13.19 & 6.43 & 16.09 & 7.54 & 16.63 & 9.19 \\
Kimi-K2-Preview & \textbf{22.92} & \textbf{16.74} & \underline{21.59} & \underline{14.36} & \textbf{25.15} & \textbf{16.20} & 26.12 & \textbf{21.86} & 20.37 & \textbf{16.81} & 19.56 & \textbf{16.01} & \textbf{24.84} & 17.00 \\
Qwen3-8B & 21.22 & 12.73 & 19.56 & 9.69 & 21.87 & 11.34 & 24.61 & 20.59 & \textbf{20.50} & 10.91 & 20.62 & 11.41 & 20.99 & 15.22 \\
Qwen3-32B & 21.81 & 14.91 & 19.99 & 11.97 & 21.98 & 14.20 & \underline{28.14} & \underline{21.77} & 19.96 & 11.42 & 19.25 & 12.99 & 22.95 & \textbf{19.61} \\
Llama-3.1-70B-Instruct & 19.09 & 9.75 & 16.16 & 8.52 & 19.03 & 10.94 & 24.13 & 11.44 & 18.99 & 8.89 & 19.71 & 9.40 & 18.39 & 9.64 \\
\bottomrule
\end{tabular}
\caption{Emotional Trajectory Volatility (ETV) measures each model's ability to promote a rapid and stable transition toward positive emotional states. Higher values suggest stronger emotional support effectiveness. For each row, the highest ETV is marked in \textbf{bold}, and the second-highest is \underline{underlined}.}
\label{ETV}
\end{table*}

\subsection{Causally-Adjusted Emotion Estimation}

\begin{figure}[t]
\centering
\includegraphics[width=0.96\columnwidth]{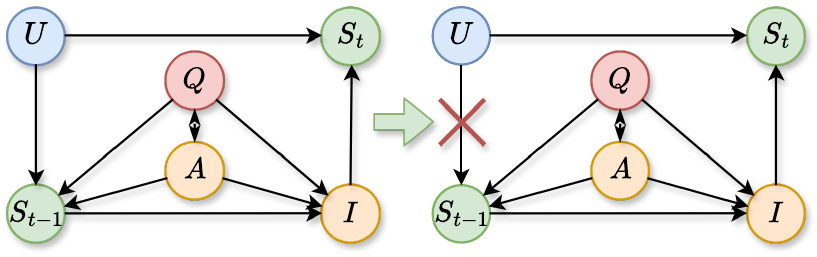} % Reduce the figure size so that it is slightly narrower than the column.
\caption{Causal graph illustrating emotional evolution with unobserved confounder. Right: theoretical intervention $\mathrm{do}(E_{t-1})$ removes spurious correlation via backdoor adjustment. Variable Definitions: $Q$ (User Dialogue History), $A$ (Model Dialogue History), $S$ (Emotion State), $I$ (Internal Thought), $U$(Unobserved Confounder).}
\label{casual graph}
\end{figure}

The effectiveness of the aforementioned metrics (BEL, ETV, ECP) hinges on accurate estimation of user emotion in each turn. Conventional models typically predict emotion from the user's immediate reply $I$ and the dialogue history ${Q,A}$, thereby estimating $P(S_t | I, Q,A)$. However, such estimation is not faithful to our proposed Markov model. Our interest lies in the direct influence between emotional states, i.e., $P(S_t | S_{t-1})$, whereas conventional approaches introduce dialogue content as an intermediary, leading to unnecessary bias. Even if $P(S_t | S_{t-1})$ could be estimated directly, spurious associations may still arise due to unobserved confounders (e.g., user personality, personal experiences) that influence both $S_{t-1}$ and $S_t$. Therefore, a shift from statistical correlation to causal effects \cite{pearl2016causal} is necessary. The ultimate goal is to estimate the post-intervention conditional probability distribution, $P(S_t|\text{do}(S_{t-1}))$, to eliminate the influence of all confounders. To this end, we construct the following causal graph (see Fig. \ref{casual graph}) to formalize the relationships between variables.

\paragraph{Core Causal Path} The dialogue history ${Q,A}$ between the user and the model influences the preceding emotional state $S_{t-1}$. Subsequently, $S_{t-1}$ and ${Q,A}$ jointly determine the user's current internal thought $I$, which in turn reflects the current emotional state $S_t$. Concurrently, we introduce an unobserved confounding variable $U$, representing stable factors that affect both preceding and current emotions. $U$ forms a backdoor path ($S_{t-1} \leftarrow U \rightarrow S_t$), which is the root cause of the spurious association between $S_{t-1}$ and $S_t$.

Based on this causal graph, we apply a theoretical intervention, $\text{do}(S_{t-1})$, to eliminate the bias from the confounder $U$. This intervention is not a practical operation but a mathematical tool. It simulates an idealized, unbiased scenario by theoretically severing all incoming paths to $S_{t-1}$ (especially the path from $U$). Using the rules of do-calculus \cite{pearl2016causal}, we transform this post-intervention distribution into a computational expression based on observable data (see the extended version for the derivation):
\begin{small}
\begin{equation}
P(S_t|\text{do}(S_{t-1}))=\mathbb{E}_{S'_{t-1},Q',A'}\mathbb{E}_{Q,A}[P(S_t|I,S'_{t-1},Q',A')].
\label{eq:causal}
\end{equation}
\end{small}

We adopt this post-intervention emotional distribution as the final, calibrated estimate of user emotion, replacing the predictions from conventional models. Based on this estimate, we construct an emotional state sequence $(s_0, s_1, \dots, s_T)$. This ensures that the subsequently computed emotion transition matrix $M$ and related metrics more faithfully reflect the model's capability to provide emotional support to the user.

\section{Experiments and Analysis}
\subsection{Evaluation Settings and Model Selection}
We sampled a total of 118 background environments for user-model interaction. For each model evaluated, we initiated dialogues based on the designed background and a first-turn exchange, which was then extended for an additional 40 turns.To mitigate the personal bias and increased time costs associated with human role-players, we adopted a methodology from existing multi-turn evaluation studies \cite{castillo2024beyond} and employed ChatGPT-4o \cite{hurst2024gpt} to simulate the user in each scenario. For each environment, we extracted 0,1 or 3 logical disturbance events, which were gradually revealed to the user-simulating model during the dialogue. For the dynamic trajectory metrics, we utilized the Skywork-Llama-3.1-Reward-v2 \cite{liu2025skywork} model to perform sentiment analysis on each dialogue turn, mapping the binary sentiment  (positive/negative) to a interval \([0, 1]\). The models selected for testing included leading closed-source models from major vendors such as OpenAI, Anthropic, Google, xAI, ByteDance, and ZhipuAI, which were accessed via their APIs. For open-source models, we selected popular and widely-used models from the community, including Phi4-14B \cite{abdin2024phi}, DeepSeek-V3 \cite{liu2024deepseek}, DeepSeek-R1 \cite{guo2025deepseek}, Qwen3-235B, Qwen3-8B, Qwen3-32B \cite{yang2025qwen3}, Mistral3.2-24B-Instruct, Kimi-K2-Preview \cite{team2025kimi}, and Llama3.1-70B-Instruct \cite{dubey2024llama}. Detailed estimation methods, evaluation specifics, parameter analysis, prompt settings, and the models' specifications and release dates are provided in the extended version, along with detailed case analyses.
\subsection{Overall Result}

Tab. \ref{BEL} and Tab. \ref{ETV} present the evaluation results for two key metrics, Baseline Emotional Level (BEL) and Emotional Trajectory Volatility (ETV), in both English and Chinese contexts. The BEL metric quantifies the user's average emotional level during multi-turn interactions, while the ETV metric measures the model's efficiency in improving a user's mood via emotional support strategies when the user is in a negative emotional state.

The BEL results reveal several key findings. First, there is no significant difference in overall emotional support capabilities between open-source and closed-source models; in fact, advanced open-source models like Kimi-K2-Preview excelled at maintaining the highest average user emotional level in multi-turn dialogues. Second, models designed specifically for role-playing did not outperform general-purpose LLMs in maintaining a user's positive emotional state. Third, models demonstrated significantly superior long-term emotional support capabilities in English compared to Chinese, as most models helped users maintain a higher average emotional level in English dialogues. Finally, regarding specific strategy application, models show a deficiency in dynamically adjusting interaction strategies based on the user's state in English; conversely, in Chinese, their application of strategies that guide users to modify their external environment to improve mood is notably weak.

The ETV results are consistent with the trends observed for BEL, while also revealing deeper dynamic features. Initially, some models that scored high on the BEL metric experienced a decline in their ETV scores. This phenomenon can be attributed to two factors: 1) when a user's mood is low, the emotional support efficiency of these models is lower than that of their counterparts; and 2) when a user's mood is positive, these models struggle to effectively maintain its stability. Therefore, a relative decline in ETV score reflects instability in a model's emotional support capability. Notably, in both English and Chinese contexts, Attention Diversion (AttDep) emerged as the most rapid emotional support strategy for improving user mood, an observation that aligns with real-world patterns.

\subsection{Emotional Centroid Position Visualization}% 情感质心位置
\begin{figure}[t]
\centering
\includegraphics[width=0.99\columnwidth]{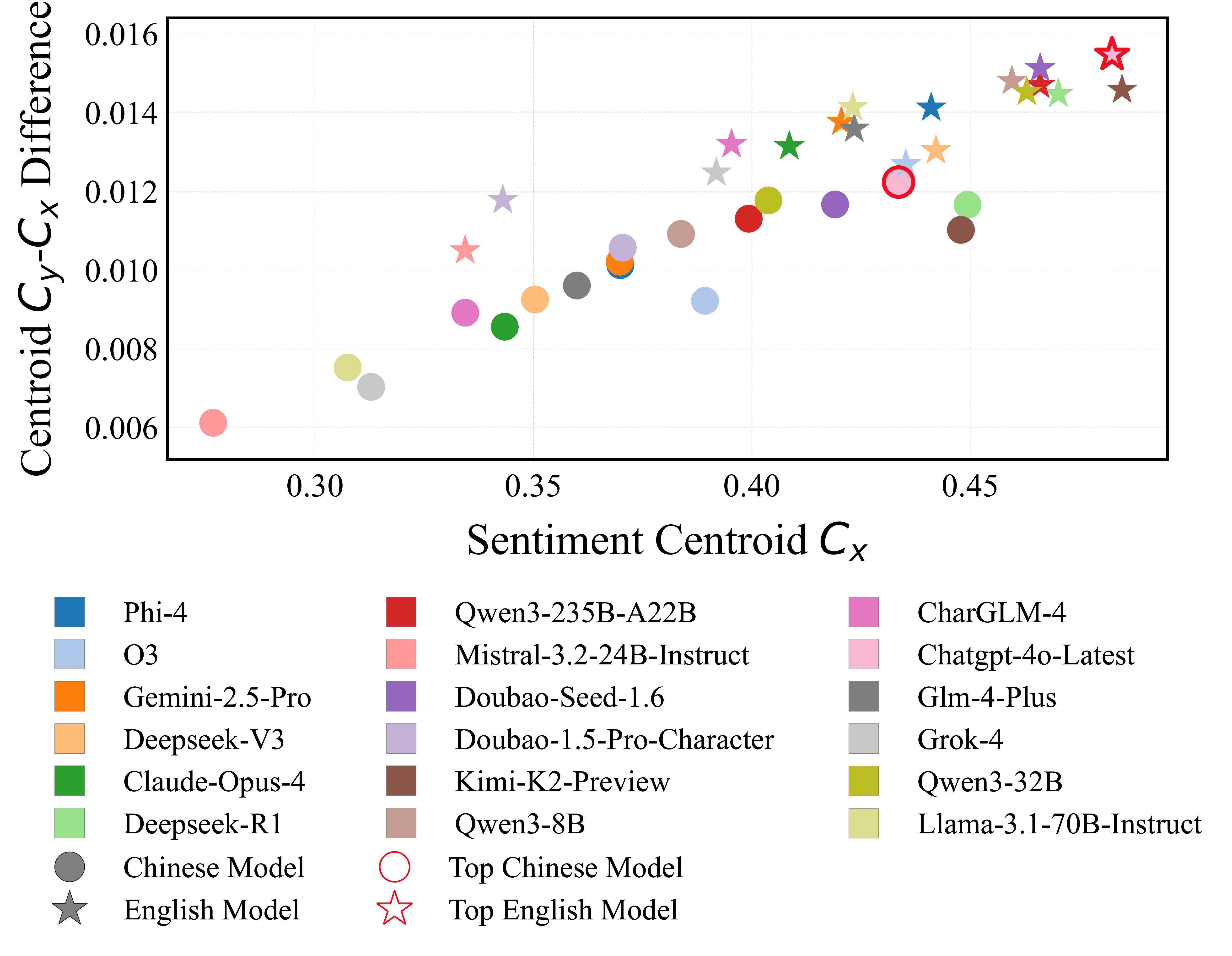} % Reduce the figure size so that it is slightly narrower than the column.
\caption{Visualization of the sentiment centroid, defined as the expected emotional position under the empirical Markov model formed by the initial distribution and transition matrix $M$.}

\label{centroids_points}
\end{figure}

We further analyze the emotional guidance capabilities of different LLMs by comparing their sentiment centroids, computed as the expected emotional position under the empirical transition model $M$. As shown in Fig. \ref{centroids_points}, the horizontal axis ($C_x$) represents the overall emotional positivity of the trajectory, while the vertical axis ($C_y - C_x$) captures the emotional concentration or consistency across turns.

\begin{figure}[ht]
\centering
\includegraphics[width=0.95\columnwidth]{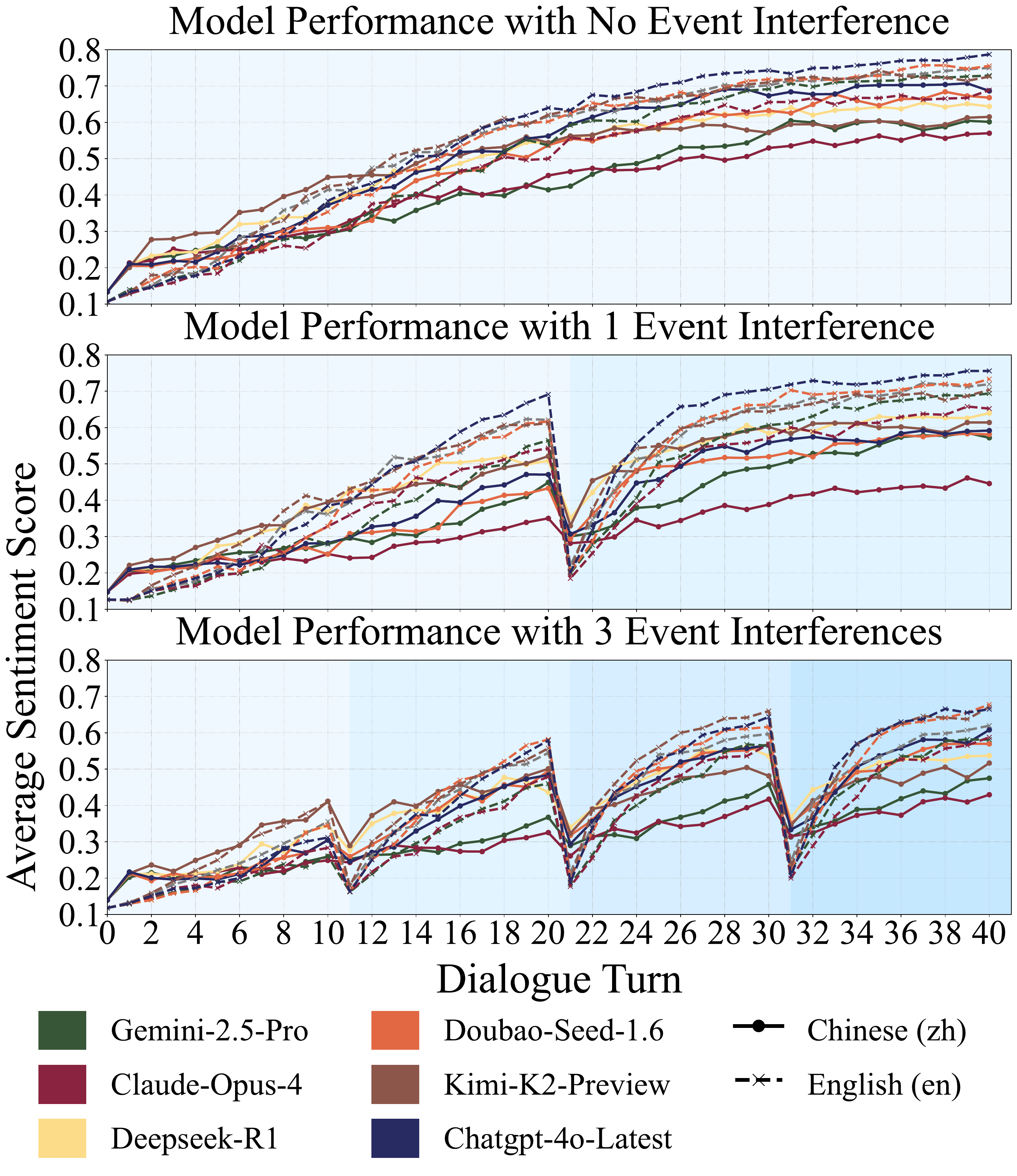} % Reduce the figure size so that it is slightly narrower than the column.
\caption{Sentiment dynamics across multi-turn dialogues under different emotional interference conditions.}
\label{emo_traj}
\end{figure}

The results reveal a clear separation between models: top-performing models, particularly those with high BEL and ETV scores, exhibit both high $C_x$ and $C_y - C_x$ values, indicating strong capabilities in steering users toward positive and stable emotional states. In contrast, models with lower centroid values either failed to maintain positive emotional progress or displayed higher volatility in user sentiment trajectories. Notably, several English-instruction-tuned models (e.g., ChatGPT-4o-Latest, kimi-K2-Preview) show superior centroid positioning compared to their Chinese counterparts, suggesting differences in emotional regulation strategies across language-specific pretraining and alignment.

\subsection{Emotional Trajectory Visualization}
% 情感轨迹图
To provide a more intuitive understanding of our proposed evaluation framework, we conduct a visual analysis under three levels of emotional disturbance (0, 1, and 3 disturbance events). For this analysis, we selected the multi-turn interaction results from 6 models. As shown in Fig. \ref{emo_traj}, the user's average emotional trajectory is tracked throughout all dialogues, where the intensity of the background shade indicates the cumulative number of disturbance events the user has experienced. The figure reveals that: (1) Models with higher ETV scores more effectively facilitate the user's emotional recovery from a low state, which corroborates our earlier assertion. (2) In the absence of disturbance events, the models can restore the user's emotion to a neutral level within a relatively short period. (3) Multiple disturbance events impede the rate of emotional recovery; however, models with stronger emotional support capabilities demonstrate greater resilience to such interferences.

\begin{table}[t]
\centering
\small
% Reduce the space between columns for this table only.
\setlength{\tabcolsep}{7pt} 
\begin{tabular}{lcccc}
% 这里有两条线
\hline
\multirow{2}{*}{\textbf{Model}} & \multicolumn{2}{c}{\textbf{ZH}} & \multicolumn{2}{c}{\textbf{EN}} \\
& \textbf{Acc(\%)} & \textbf{$\Delta$} & \textbf{Acc(\%)} & \textbf{$\Delta$} \\
\hline
Chatgpt-4o & \textbf{76.57} & {-} & \textbf{90.91} & {-} \\
Deepseek-V3 & 73.20 & {-} & 89.47 & {-} \\
RRM-7B & 72.96 & {-} & 87.44 & {-} \\
Llama3.1-8B-IT. & 74.76 & {-} & 88.88 & {-} \\
\hline
\multicolumn{5}{l}{\textit{Skywork-Reward-Llama-3.1-B-v0.2}} \\
\hline
\quad \textit{w/o} CA & 73.82 & \textit{ref} & 89.11 & \textit{ref} \\
\quad \textit{w/} CA & 74.61 & 0.79$\uparrow$ & 89.59 & 0.48$\uparrow$ \\
\hline
\multicolumn{5}{l}{\textit{Skywork-Reward-V2-Llama-3.1-8B}} \\
\hline
\quad \textit{w/o} CA & 72.18 & \textit{ref} & 89.59 & \textit{ref} \\
\quad \textit{w/} CA & \underline{74.92} & 2.74$\uparrow$ & \underline{89.83} & 0.24$\uparrow$ \\
\hline
\end{tabular}
\caption{Performance comparison on the human-annotated dataset, with ablation results showing the improvement ($\Delta$) from our causally-adjusted (CA) estimation method. For each case, the model with the highest Acc is marked in \textbf{bold}, and the second-highest is \underline{underlined}.}
\label{ablation}

\end{table}

\subsection{Causal-Enhanced Estimation of Emotional Dynamics} 
% 因果推断对于情感评估的效果

To evaluate the consistency of our sentiment recognition model with human perception and validate our estimation calibration method, we constructed a human-annotated multi-turn dialogue dataset. This dataset comprises nearly 2,000 Chinese and English multi-turn dialogues selected from the Daily Dialog and CPED corpora. Three expert annotators were employed to re-label the sentiment of each turn in these dialogues with binary annotations, building upon the existing labels. Details can be found in the extended version. As shown in Tab. \ref{ablation}, our findings fall into two aspects. Firstly, comparative experiments demonstrate that our proposed estimation calibration method effectively enhances the model's sentiment recognition capabilities by mitigating the influence of confounding factors. The application of our unbiased estimation method led to improved sentiment recognition performance across different models. Notably, our approach achieves state-of-the-art performance compared to other existing models. Secondly, our evaluation model, combined with the calibration method, shows a high degree of consistency with human judgments, reaching an accuracy of 75\% on Chinese dialogues and 90\% on English dialogues.

\section{Conclusions}
In this paper, we propose a dynamic trajectory analysis framework to evaluate the emotional support capabilities of language models. The core of this framework simulates the real user-model interaction process, guiding the interaction by constructing background contexts, introducing multi-strategy constraints, and incorporating event-driven perturbations. We have designed metrics for dynamic trajectory analysis from three perspectives and utilize causal inference to calibrate the evaluation outcomes. Experiments confirm that our methodology offers a more comprehensive and multifaceted assessment of a model's emotional support capabilities, showing high consistency with human evaluations.

\bibliography{aaai2026}

\appendix
\newpage\onecolumn

\section{A. Benchmark Design and Data Construction}

\subsection{A.1 Dataset Details and Statistics}
\subsubsection{User-Side}From the user’s perspective, we design emotionally challenging scenarios to simulate real-life contexts where emotional regulation is critical. We categorize emotionally suppressed situations into four major domains. Each domain encompasses multiple prototypical triggers of emotional distress:

\begin{table}[ht]
\centering
\begin{tabular}{p{3.5cm}p{4.5cm}p{7cm}}
\hline
\textbf{Category} & \textbf{Subcategory} & \textbf{Description} \\
\hline
\multirow{4}{*}{Professional and Social} 
& Occupational Stress & Emotional strain resulting from work overload, job pressure, or workplace burnout. \\
& Interpersonal Conflict & Tensions or disputes in social or professional relationships. \\
& Career Development and Insecurity & Anxiety about future prospects, job stability, or professional growth. \\
& Social Comparison and Identity & Stress arising from comparing oneself to others or facing identity struggles in social contexts. \\
\hline

\multirow{3}{*}{Intimate Relationships}
& Romantic Relationships & Emotional issues involving love, breakups, or commitment difficulties. \\
& Family and Kinship & Conflicts or emotional stress within family or close kin relations. \\
& Friendships & Tensions, distance, or misunderstandings in platonic friendships. \\
\hline

\multirow{4}{*}{\parbox[c]{3.5cm}{Personal Growth and\\Internal Conflict}}
& Academic and Developmental Stress & Performance pressure related to school, learning, or future expectations. \\
& Self-perception and Sense of Self-worth & Internal struggles with self-image, confidence, or identity. \\
& Psychosomatic Health Issues & Emotional distress manifesting through physical symptoms such as fatigue or insomnia. \\
& Existential and Meaning Crisis & Deep questioning about life purpose, meaning, or personal direction. \\
\hline

\multirow{3}{*}{\parbox[c]{3.5cm}{Life Circumstances and\\Acute Events}}
& Economic and Subsistence Pressure & Financial hardship or concerns about basic survival and resources. \\
& Major Negative Life Events & Emotional impact of events such as bereavement, accidents, or serious loss. \\
& Environmental and Social Factors & Distress linked to surroundings, societal instability, or discrimination. \\
\hline
\end{tabular}
\caption{categorization of emotional distress types across common life domains}
\label{tab:emotion-contexts}
\end{table}

\begin{figure*}[ht]
\centering
\includegraphics[width=0.9\textwidth]{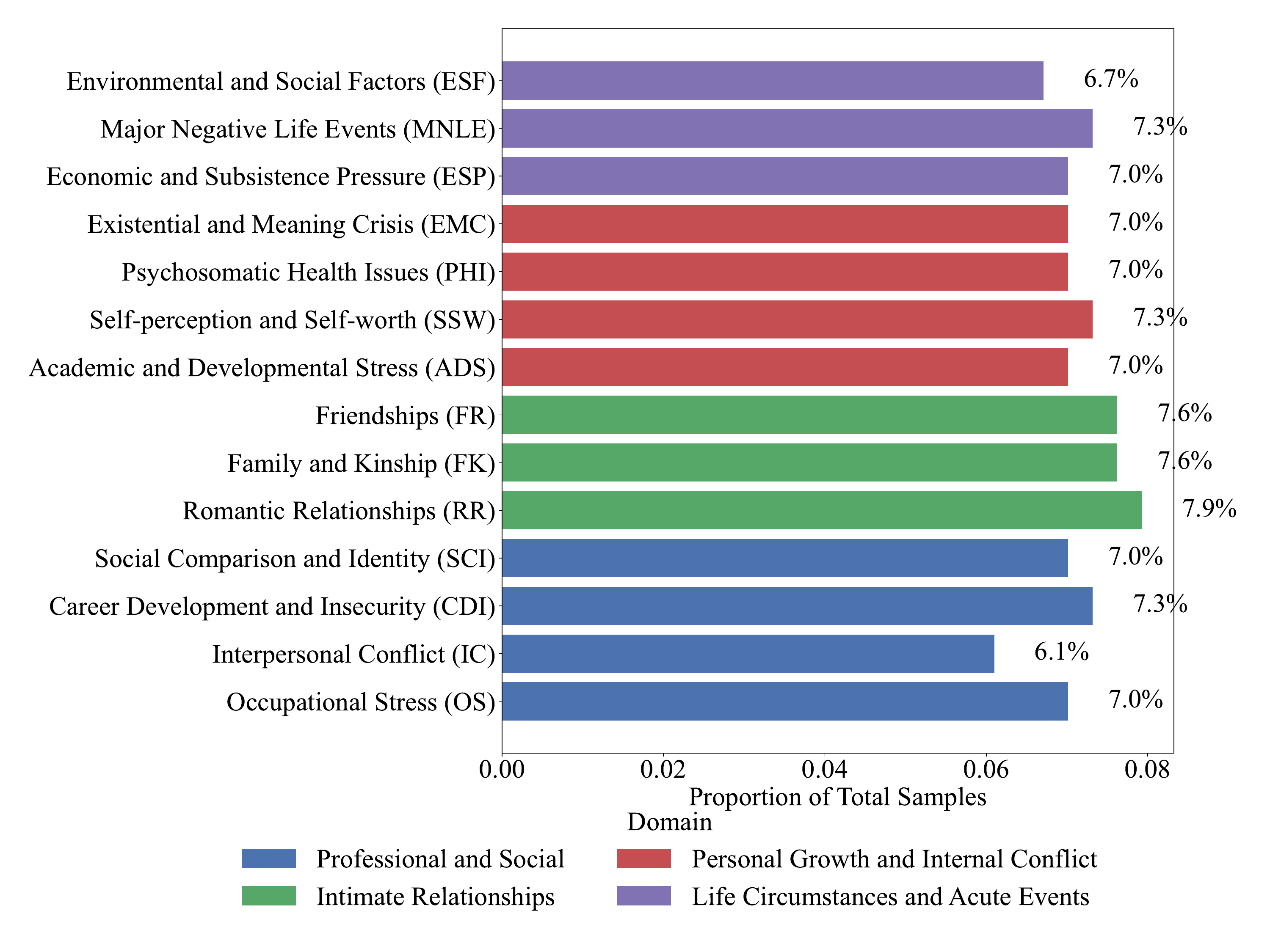} % Reduce the figure size so that it is slightly narrower than the column.
\caption{distribution of user emotional distress scenarios}
\label{scenario}
\end{figure*}

As illustrated in Fig.~\ref{scenario}, to ensure balanced coverage of emotionally salient scenarios, we manually constructed the dataset by assigning a comparable number of instances to each subcategory. Within each category, we selected 3–4 representative subcategories grounded in psychological literature and commonly observed user distress themes. The sample counts per subcategory were intentionally designed to maintain distributional balance, reduce data-induced biases during evaluation, and ensure that all types of emotional challenges are equally represented during model testing.

\subsubsection{Model-Side}

\begin{figure*}[ht]
\centering
\includegraphics[width=0.8\textwidth]{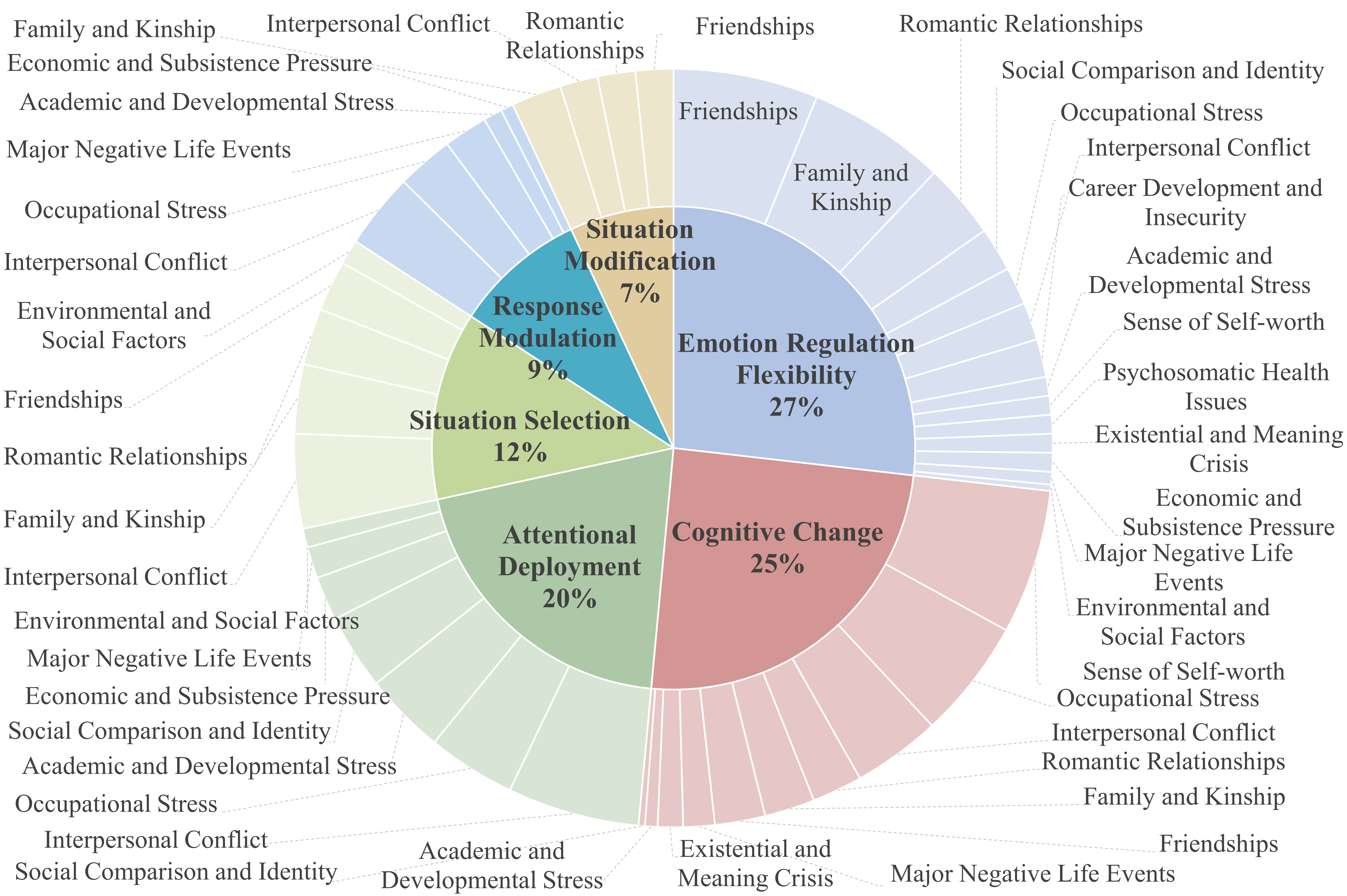} % Reduce the figure size so that it is slightly narrower than the column.
\caption{Distribution of emotion regulation strategies guiding model responses, constrained by mainstream psychological theory to promote user emotional improvement.}
\label{data dist}
\end{figure*}

As illustrated in Figure~\ref{data dist}, our data construction systematically covers diverse emotion scenarios and regulatory strategies. Each emotional context is associated with multiple regulation dimensions. However, not all strategies are equally applicable across emotion types. For instance, in scenarios related to \textit{self-perception and self-worth}, \textbf{Cognitive Change} tends to be more effective, as negative emotions often stem from internal cognitive biases, making environment-focused strategies like Situation Selection or Modification less impactful. To account for strategy-context suitability, we proportionally adjusted the sampling distribution to assign higher weights to strategies that are more contextually appropriate. This design ensures that our evaluation not only reflects the diversity of regulation strategies but also their actual effectiveness in context, thereby enhancing the psychological validity and discriminative power of the benchmark. It avoids simplistic metrics based on strategy count and instead emphasizes meaningful regulatory impact.

\subsection{A.2 Data Quality Assurance}
We construct a total of 328 emotional distress scenarios, spanning a wide range of social and individual psychological domains. To simulate emotional decline, we design 1,152 disturbance events that can further aggravate the user’s emotional state. All generated data undergoes meticulous human inspection by a dedicated team of quality reviewers. We ensure that each dialogue exhibits natural emotional expression, coherent logic, and consistency with the annotated emotion regulation mechanisms. Any data identified as low-quality is excluded to maintain the reliability and credibility of the evaluation framework:

\begin{itemize}
 \item \textbf{Role-Immersive Annotator Training.}
 We design an immersive training protocol in which annotators simulate both user and assistant roles across emotional support dialogues. Trainees complete guided exercises covering typical error cases (e.g., inconsistent emotion labels, mismatched tone) and must pass a qualification test before production annotation begins.
\item \textbf{Redundancy with Psychological Consistency Checks.}
 Each dialogue is annotated independently by two reviewers. Beyond surface-level agreement, we introduce consistency constraints based on psychological plausibility—for example, ensuring that emotional transitions match known response patterns. Samples failing this check are re-reviewed by a senior annotator.
\item \textbf{Contextual Drift Auditing.}
 We monitor whether annotators maintain coherence across multi-turn conversations, especially in long interaction chains. A special check focuses on "drift" errors—where emotional labels lose alignment with the evolving narrative. These samples are flagged for correction and used as feedback material in team debriefs.
\item \textbf{Dynamic Quality Calibration.}
 We divide annotation into rolling batches and perform retrospective calibration checks every 200 samples. Annotators receive feedback dashboards showing inter-annotator agreement trends, correction frequencies, and breakdowns by emotion category. Annotators with declining consistency are re-trained or paused.
\item \textbf{Expert-In-The-Loop Final Validation.}
 All high-impact or ambiguous cases—such as borderline emotional states or conflicting triggers—are escalated to expert reviewers with backgrounds in clinical psychology or affective computing. These experts provide final adjudications and maintain a reference log to guide future judgments.
 \end{itemize}

\section{B. Theoretical Analysis and Proofs}
\subsection{B.1 Formal Proof of Estimating the Transition Matrix via MLE}

The emotional transition matrix $M$ is an $N \times N$ matrix, where each entry $m_{i,j}$ denotes the conditional probability of transitioning from emotional state $e_i$ to $e_j$ after one round of interaction with the model: $m_{i,j} = P(s_t=e_j | s_{t-1}=e_i), \quad \forall e_i,e_j \in \mathcal{S}$. Specifically, for each predefined evaluation environment, we simulate a complete dialogue consisting of $T$ interaction turns. This yields an observed emotional state trajectory $(s_0, s_1, \dots, s_T)$, where $s_t \in \mathcal{E}$. Based on this trajectory, we estimate the emotional transition matrix $M$ corresponding to that context. Firstly, We construct a transition count matrix $F=[f_{ij}] \in \mathbb{Z}^{N \times N}$ where:
\begin{equation}
f_{ij} = \left| \left\{ t \mid s_{t-1} = e_i \land s_t = e_j, \; 1 \leq t \leq T \right\} \right|.
\end{equation}
Here, $|\cdot|$ denotes the cardinality of a set. We then normalize each row of the frequency matrix to obtain the transition probabilities $m_{ij}$ as the Maximum Likelihood Estimate (MLE). This estimated value represents the empirical probability that, within the given dialogue, the user transitions from state $e_i$ to state $e_j$ after one model interaction:
\begin{equation}
m_{ij} = \frac{f_{ij}}{\sum_{k=1}^{N} f_{ik}} \quad \text{if } \sum_k f_{ik} > 0.
\end{equation}

This formulation is defined only when the denominator satisfies $\sum_{k=1}^{N} f_{ik} > 0$, meaning that the state $s_i$ must occur at least once as a source state in the dialogue.

\paragraph{Theorem.}
For a sequence of observations generated by a first-order Markov process, the empirical transition matrix $M$ estimated from the sample sequence is the unique maximum likelihood estimator (MLE) of the true transition probabilities.

\paragraph{Proof.}
Assume that the observed sequence $(s_0, s_1, \dots, s_T)$ is generated by an unknown, time-homogeneous first-order Markov process with true transition probabilities $p_{ij} = P(s_t = e_j \mid s_{t-1} = e_i)$. Conditional on the initial state $s_0$, the likelihood function of observing the entire sequence is given by:
\begin{equation}
    \mathcal{L}(P) = \prod_{t=1}^{T} P(s_t \mid s_{t-1}) = \prod_{i=1}^{N} \prod_{j=1}^{N} p_{ij}^{f_{ij}},
\end{equation}
where $P = [p_{ij}]$ is the transition matrix to be estimated, and $f_{ij}$ denotes the number of observed transitions from state $s_i$ to $s_j$. Our goal is to find the transition probabilities $p_{ij}$ that maximize $\mathcal{L}(P)$, subject to the constraint $\sum_{j=1}^{N} p_{ij} = 1$ for all $i$. We maximize the log-likelihood function:
\begin{equation}
\log \mathcal{L}(P) = \sum_{i=1}^{N} \sum_{j=1}^{N} f_{ij} \log p_{ij} ,   
\end{equation}
according to the method of Lagrange multipliers, we construct the $ \Lambda(P, \lambda)$:
\begin{equation}
    \Lambda(P, \lambda) = \sum_{i=1}^{N} \sum_{j=1}^{N} f_{ij} \log p_{ij} + \sum_{i=1}^{N} \lambda_i \left(1 - \sum_{j=1}^{N} p_{ij} \right).
\end{equation}
Taking partial derivatives with respect to $p_{ij}$ and setting to zero:
\begin{equation}
    \frac{\partial \Lambda}{\partial p_{ij}} = \frac{f_{ij}}{p_{ij}} - \lambda_i = 0 \Rightarrow p_{ij} = \frac{f_{ij}}{\lambda_i},
\end{equation}
substituting into the normalization constraint:
\begin{equation}
\sum_{j=1}^{N} \frac{f_{ij}}{\lambda_i} = 1 \Rightarrow \frac{1}{\lambda_i} \sum_{j=1}^{N} f_{ij} = 1 \Rightarrow \lambda_i = \sum_{j=1}^{N} f_{ij},
\end{equation}
plugging back into the expression for $P_{ij}$:
\begin{equation}
\hat{p}_{ij} = m_{ij} = \frac{f_{ij}}{\sum_{k=1}^{N} f_{ik}}.
\end{equation}

This completes the proof that the empirical matrix $M$ we construct is not an arbitrary design, but rather the statistically optimal estimate of the underlying Markov process governing the observed sequence.

\subsection{B.2 Theoretical Justification of the ECP}

\paragraph{Theorem.}
\textbf{Emotional Centroid Position (ECP)} is a probabilistic measure defined by the empirical transition probability matrix $M$ and the associated initial state distribution (computed from our estimated $M$), and is numerically equivalent to the centroid of the corresponding empirical transition distribution in the emotional state space (i.e., the average observed emotional position):

\textbf{Definition 1 (Empirical Centroid of Transition Distribution).} 
Suppose we \textbf{observe} $T$ state transitions forming a dataset $\{(s_{t-1}, s_t)\}_{t=1}^T$, where $s_t \in \{e_1, \dots, e_N\}$ represents discrete emotional states. The empirical centroid $C_{\text{emp}} = (\bar{s}_x, \bar{s}_y)$ is defined as the arithmetic mean of the coordinates:
\begin{equation}
\bar{s}_x = \frac{1}{T} \sum_{t=1}^{T} s_{t-1}, \quad \bar{s}_y = \frac{1}{T} \sum_{t=1}^{T} s_t.
\end{equation}
\textbf{Definition 2 (Emotional Centroid Position, ECP).}  
As defined in the main text, the ECP is the expectation of the transition pair:
\begin{equation}
C_{\mathcal{E}} = (E[s_{t-1}], E[s_t]).
\end{equation}
This expectation depends on the joint probability $P(s_{t-1} = e_i, s_t = e_j)$, which is derived from the empirical statistics of the observed sequence:
\begin{equation}
P(s_{t-1}=e_i, s_t=e_j) = P(s_t = e_j \mid s_{t-1} = e_i) \cdot P(s_{t-1} = e_i).
\end{equation}
Here, $P(s_t = e_j \mid s_{t-1} = e_i) = m_{ij}$ is the element of the empirical transition matrix $M$, and $P(s_{t-1} = e_i) = N_i/T$ is the empirical marginal distribution, where $N_i$ is the number of times $e_i$ appears as a source state.

\paragraph{Proof.}
\textbf{Step 1: X-coordinates Equivalence ($E[s_{t-1}] = \bar{s}_x$)}  
According to the definition of expectation:
\begin{equation}
E[s_{t-1}] = \sum_{i=1}^{N} e_i \cdot P(s_{t-1} = e_i) = \sum_{i=1}^{N} e_i \cdot \frac{N_i}{T}.
\end{equation}

The summation $\sum_{i=1}^{N} e_i \cdot N_i$ represents the total obtained by multiplying each initial emotional state $e_i$ by its frequency of occurrence $N_i$, and then summing the results. Numerically, this is equivalent to directly summing all the initial emotional states over the $T$ transitions: $\sum_{t=1}^{T} s_{t-1}$. Therefore:
\begin{equation}
E[s_{t-1}] = \frac{1}{T} \sum_{t=1}^{T} s_{t-1} = \bar{s}_x.
\end{equation}

\textbf{Step 2: Y-coordinates Equivalence ($E[s_t] = \bar{s}_y$)}  
According to the definition of expectation and the joint probability distribution:
\begin{equation}
E[s_t] = \sum_{i=1}^{N} \sum_{j=1}^{N} e_j \cdot P(s_{t-1} = e_i, s_t = e_j)
= \sum_{i=1}^{N} \sum_{j=1}^{N} e_j \cdot m_{ij} \cdot P(s_{t-1} = e_i).
\end{equation}

Substituting the empirical estimate of the transition matrix $m_{ij} = \frac{f_{ij}}{N_i}$ (where $f_{ij}$ is the count of transitions from $e_i$ to $e_j$), and the marginal distribution $P(s_{t-1} = e_i) = \frac{N_i}{T}$:
\begin{equation}
E[s_t] = \sum_{i=1}^{N} \sum_{j=1}^{N} e_j \cdot \left( \frac{f_{ij}}{N_i} \right) \cdot \left( \frac{N_i}{T} \right).
\end{equation}

In the summation, for all terms where $N_i > 0$, the $N_i$ terms cancel out. For $N_i = 0$, the corresponding $f_{ij} = 0$, so those terms vanish. Therefore, the equation simplifies to:
\begin{equation}
E[s_t] = \frac{1}{T} \sum_{i=1}^{N} \sum_{j=1}^{N} e_j \cdot f_{ij}.
\end{equation}

The summation $\sum_{i,j} e_j \cdot f_{ij}$ represents the total obtained by multiplying each target emotional value $e_j$ by its total number of occurrences (i.e., $\sum_i f_{ij}$), and then summing the results. Numerically, this is equivalent to directly summing all target emotional values over the $T$ transitions: $\sum_{t=1}^{T} s_t$. Therefore:
\begin{equation}
E[s_t] = \frac{1}{T} \sum_{t=1}^{T} s_t = \bar{s}_y.
\end{equation}

\textbf{Conclusion.}  
We have shown that: $C_{\mathcal{E}} = (E[s_{t-1}], E[s_t]) = (\bar{s}_x, \bar{s}_y) = C_{\text{emp}}$. This proves that the Emotional Centroid Position, as derived from the empirical transition matrix $M$, is numerically identical to the centroid of the empirical transition distribution in the emotional state space.

\subsection{B.3 Causally-Adjusted Emotion Estimation}

\subsubsection{Preliminaries.}

Judea Pearl's framework \cite{pearl2016causal} of causal inference provides a formal mathematical foundation for distinguishing causation from correlation. It employs Structural Causal Models (SCMs), often visualized as Directed Acyclic Graphs (DAGs), to encode assumptions about the underlying data-generating processes. A central concept in this framework is the \textit{do}-operator, denoted as $\text{do}(X=x)$, which simulates a physical intervention by forcing a variable $X$ to take on a value $x$. This allows for the rigorous definition and identification of causal effects. To determine if a causal effect is identifiable from observational data, the framework provides a set of powerful tools, which we outline below.

\textbf{Back-Door Criterion.} Given an ordered pair of variables $(X, Y)$ in a DAG, a set of variables $Z$ satisfies the Back-Door Criterion if no node in $Z$ is a descendant of $X$, and $Z$ blocks every path between $X$ and $Y$ that contains an arrow into $X$. If $Z$ satisfies this criterion, the causal effect is identifiable by the adjustment formula:
\begin{equation}
    P(Y | \text{do}(X=x)) = \sum_z P(Y | X=x, Z=z) P(Z=z).
\end{equation}

\textbf{Front-Door Criterion.} A set of variables $Z$ satisfies the Front-Door Criterion relative to $(X, Y)$ if: (1) $Z$ intercepts all directed paths from $X$ to $Y$; (2) there is no unblocked back-door path from $X$ to $Z$; and (3) all back-door paths from $Z$ to $Y$ are blocked by $X$. If $Z$ satisfies this criterion and $P(x, z) > 0$, the causal effect is identifiable and given by:
\begin{equation}
    P(Y | \text{do}(X=x)) = \sum_z P(Z=z | X=x) \sum_{x'} P(Y | X=x', Z=z) P(X=x').
\end{equation}

\textbf{The Rules of Do-calculus.} Do-calculus offers a complete system for symbolic manipulation of interventional distributions. Let $X, Y, Z,$ and $W$ be disjoint sets of nodes in a causal DAG $G$. The three fundamental rules are as follows:

% \textbf{Rule 1: Insertion/Deletion of Observation.} This rule states that an observation can be added or removed from the set of conditioning variables.
% \begin{equation}
%     P(y | \text{do}(x), z, w) = P(y | \text{do}(x), w) \quad \text{if } (Y \perp Z | X, W)_{G_{\bar{X}}}.
% \end{equation}
% The condition holds if the set of variables $Y$ is independent of $Z$ given $X$ and $W$ in the graph $G_{\bar{X}}$, which is the graph $G$ with all incoming arrows to nodes in $X$ removed.

% \textbf{Rule 2: Action/Observation Exchange.} This rule allows for the exchange of an intervention with an observation.
% \begin{equation}
%     P(y | \text{do}(x), \text{do}(z), w) = P(y | \text{do}(x), z, w) \quad \text{if } (Y \perp Z | X, W)_{G_{\bar{X}\underline{Z}}}.
% \end{equation}
% The condition holds if $Y$ is independent of $Z$ given $X$ and $W$ in the graph $G_{\bar{X}\underline{Z}}$, which is derived from $G$ by removing all arrows pointing into $X$ and all arrows pointing out of $Z$.

% \textbf{Rule 3: Insertion/Deletion of Action.} This rule allows for an entire intervention to be added or removed.
% \begin{equation}
%     P(y | \text{do}(x), \text{do}(z), w) = P(y | \text{do}(x), w) \quad \text{if } (Y \perp Z | X, W)_{G_{\bar{X}, \overline{Z(W)}}}.
% \end{equation}
% The condition holds if $Y$ is independent of $Z$ given $X$ and $W$ in the graph $G_{\bar{X}, \overline{Z(W)}}$, where this graph is derived from $G$ by removing all arrows pointing into $X$ and all arrows pointing out of $Z$ for those nodes in $Z$ that are not ancestors of any node in $W$ in the graph $G_{\bar{X}}$.

\textbf{Rule 1: Deletion of Observation.} This rule states that an observation can be added or removed from the set of conditioning variables.
\begin{equation}
    P(Y | \text{do}(X), Z, W) = P(Y | \text{do}(X), W) \quad \text{if } (Y \perp Z | X, W)_{G_{\bar{X}}}.
\end{equation}
The condition holds if the set of variables $Y$ is independent of $Z$ given $X$ and $W$ in the graph $G_{\bar{X}}$, which is the graph $G$ with all incoming arrows to nodes in $X$ removed.

\textbf{Rule 2: Action/Observation Exchange.} This rule allows for the exchange of an intervention with an observation.
\begin{equation}
    P(Y | \text{do}(X), \text{do}(Z), W) = P(Y | \text{do}(X), Z, W) \quad \text{if } (Y \perp Z | X, W)_{G_{\bar{X}\underline{Z}}}.
\end{equation}
The condition holds if $Y$ is independent of $Z$ given $X$ and $W$ in the graph $G_{\bar{X}\underline{Z}}$, which is derived from $G$ by removing all arrows pointing into $X$ and all arrows pointing out of $Z$.

\textbf{Rule 3: Ignoring Action.} This rule allows for an entire intervention to be added or removed.
\begin{equation}
    P(Y | \text{do}(X), \text{do}(Z), W) = P(Y | \text{do}(X), W) \quad \text{if } (Y \perp Z | X, W)_{G_{\bar{X}, \overline{Z(W)}}}.
\end{equation}
The condition holds if $Y$ is independent of $Z$ given $X$ and $W$ in the graph $G_{\bar{X}, \overline{Z(W)}}$, where this graph is derived from $G$ by removing all arrows pointing into $X$ and all arrows pointing out of $Z$ for those nodes in $Z$ that are not ancestors of any node in $W$ in the graph $G_{\bar{X}}$.

\subsubsection{Proof of Identifiability for the Causal Emotion Effect.}

Armed with these foundational principles, we can now apply them to the specific causal problem addressed in this paper. Under the assumptions of the causal graph constructed in the main text (Fig. \ref{casual graph}), the causal effect of the user's emotional state in the previous turn, $S_{t-1}$, on the current emotional state, $S_t$, is identifiable. This implies that, despite the existence of unobserved confounders $U$, the interventional distribution $P(E_t\,|\,\text{do}(S_{t-1}))$ can be fully expressed in terms of probability distributions estimable from observational data. Specifically, this causal effect can be computed by adjusting for relevant paths in the causal graph, leading to the following estimation formula:
\begin{equation}
    P(S_t|\text{do}(S_{t-1}))=\mathbb{E}_{S'_{t-1},Q',A'}\mathbb{E}_{Q,A}[P(S_t|I,S'_{t-1},Q',A')]
\end{equation}

\paragraph{Proof.}
To disentangle confounding factors from the emotional prediction process in the causal graph, we apply rules from do-calculus along with other causal inference techniques to derive an unbiased estimation of emotional states. Our focus lies in assessing the causal impact of the emotional state at the previous time step on the current emotional state. The causal intervention process can be formally described by the following mathematical formulation:
\begin{align}
P(S_t\,|\,\text{do}(S_{t-1})) 
&= \sum_{I} P(S_t\,|\,\text{do}(S_{t-1}), I)\, P(I\,|\,\text{do}(S_{t-1})) \nonumber\\
&\overset{\text{\textit{Rule 2}}}{=} \sum_{I} P(S_t\,|\,\text{do}(S_{t-1}), \text{do}(I))\, P(I\,|\,\text{do}(S_{t-1})) \nonumber\\
&\overset{\text{\textit{Backdoor}}}{=} \sum_{I} P(S_t\,|\,\text{do}(S_{t-1}), \text{do}(I)) \sum_{Q, A} P(I\,|\,S_{t-1}, Q, A)\, P(Q, A) \nonumber\\
&\overset{\text{\textit{Rule 3}}}{=} \sum_{I} P(S_t\,|\,\text{do}(I)) \sum_{Q, A} P(I\,|\,S_{t-1}, Q, A)\, P(Q, A) \nonumber\\
&= \sum_{I} \sum_{S_{t-1}', Q', A'} P(S_t\,|\,\text{do}(I), S_{t-1}', Q', A')\, P(S_{t-1}', Q', A'\,|\,\text{do}(I)) \sum_{Q, A} P(I\,|\,S_{t-1}, Q, A)\, P(Q, A) \nonumber\\
&\overset{\text{\textit{Rule 3}}}{=} \sum_{I} \sum_{S_{t-1}', Q', A'} P(S_t\,|\,\text{do}(I), S_{t-1}', Q', A')\, P(S_{t-1}', Q', A') \sum_{Q, A} P(I\,|\,S_{t-1}, Q, A)\, P(Q, A) \nonumber\\
&\overset{\text{\textit{Rule 2}}}{=} \sum_{I} \sum_{S_{t-1}', Q', A'} P(S_t\,|\,I, S_{t-1}', Q', A')\, P(S_{t-1}', Q', A') \sum_{Q, A} P(I\,|\,S_{t-1}, Q, A)\, P(Q, A) \nonumber\\
&= \sum_{I} \sum_{S_{t-1}', Q', A'} \sum_{Q, A} P(S_t\,|\,I, S_{t-1}', Q', A')\, P(S_{t-1}', Q', A')\, P(I\,|\,S_{t-1}, Q, A)\, P(Q, A).
\end{align}
Here, $S'_{t-1}, Q', A'$ are summation variables over their respective state spaces, used to compute expectations. They are distinguished from the existing variables $S_{t-1}, Q, A$ inside the expression to avoid confusion. Based on this, the above equation can be rewritten in the form of an expectation:
\begin{align}
P(S_t\,|\,\text{do}(S_{t-1})) 
&= \sum_{I} \sum_{S_{t-1}', Q', A'} \sum_{Q, A} P(S_t\,|\,I, S_{t-1}', Q', A')\, P(S_{t-1}', Q', A')\, P(I\,|\,S_{t-1}, Q, A)\, P(Q, A) \nonumber\\
&= \sum_{S_{t-1}', Q', A'} \left[ \sum_{I} \sum_{Q, A} P(S_t\,|\,I, S_{t-1}', Q', A')\, P(I\,|\,S_{t-1}, Q, A)\, P(Q, A) \right] P(S_{t-1}', Q', A') \nonumber\\
&= \mathbb{E}_{S_{t-1}', Q', A'} \left[ \sum_{Q, A} \left[ \sum_{I} P(S_t\,|\,I, S_{t-1}', Q', A')\, P(I\,|\,S_{t-1}, Q, A) \right] P(Q, A) \right] \nonumber\\
&= \mathbb{E}_{S_{t-1}', Q', A'}\, \mathbb{E}_{Q, A} \left[ \sum_{I} P(S_t\,|\,I, S_{t-1}', Q', A')\, P(I\,|\,S_{t-1}, Q, A) \right].
\end{align}
Where $I$ is defined as the user's internal thought or communicative intent, which represents a cognitive product formed after receiving information but before generating a concrete verbal response. This internal thought serves as the underlying driver of both emotional change and language production.

Based on this definition, we introduce a determinism assumption regarding the user’s cognitive process: given a specific prior emotional state $S_{t-1} = e^\star$ and a fixed dialogue history $(Q = q^\star, A = a^\star)$, the user's internal thought $I$ is uniquely determined and denoted as $i^\star$.

This assumption posits that, although the surface-level verbal expressions may exhibit minor variations, the core reasoning process underlying these responses remains constant for the same user under identical historical conditions. In probabilistic terms, this assumption implies that the conditional probability $P(I|,S_{t-1} = e^\star, Q = q^\star, A = a^\star)$ is a Kronecker delta function centered at $I = i^\star$, i.e., it equals 1 when $I = i^\star$ and 0 otherwise.

Under this assumption, the summation over $I$ in the original expression can be simplified. Accordingly, the equation becomes:

\begin{align}
P(S_t\,|\,\text{do}(S_{t-1})) 
&= \mathbb{E}_{S_{t-1}', Q', A'}\, \mathbb{E}_{Q, A} \left[ \sum_{I} P(S_t\,|\,I, S_{t-1}', Q', A')\, P(I\,|\,S_{t-1}, Q, A) \right]  \nonumber\\
% &= \mathbb{E}_{S_{t-1}', Q', A'}\, \mathbb{E}_{Q, A} \left[ P(S_t\,|\,I^{\star}_{S_{t-1}^{\star}, Q^{\star}, A^{\star}}, S_{t-1}', Q', A') \right]  \nonumber\\
&= \mathbb{E}_{S_{t-1}', Q', A'}\, \mathbb{E}_{Q, A} \left[ P(S_t\,|\,I, S_{t-1}', Q', A') \right].
\end{align}

\section{C. Experimental Setup and Implementation Details}
This section will introduce the methodology, experimental details, and specific experimental parameters, respectively.

\subsection{C.1 Code and Data.} 
% We have released \textbf{part of our evaluation method's code and some of the test data in the supplementary materials}. We will open-source all of the code and data to the public after the paper is published.

\begin{tcolorbox}[colback=gray!5, colframe=black!80, title=\textbf{Release Statement}]
\textbf{Code:} The current submission includes the \textbf{User and Model environments}, \textbf{dynamic interaction scripts}, and relevant \textbf{evaluation code}. Upon public release of this paper, we will open-source the complete evaluation framework.

\vspace{1mm}
\textbf{Data:} We currently provide \textbf{60 test samples}. After the paper is made publicly available, we will release the \textbf{full dataset} to support reproducibility and further research.
\end{tcolorbox}

\subsection{C.2 Methodology and Experimental Details}

\subsubsection{Emotional Trajectory Volatility (ETV).} The Emotional Trajectory Volatility (ETV) is designed to measure an inherent asymmetric capability to guide user emotions towards a positive direction and inhibit regressions to negative states. The theoretical form of this metric is defined as:
\begin{equation}
ETV = \sum_{1 \le i < j \le N} \omega(e_i) \cdot (e_j - e_i) \cdot (m_{i,j} - m_{j,i}).
\end{equation}
Here, the transition probability $m_{i,j}$ corresponds to the probability $P(s_t=e_j|s_{t-1}=e_i)$ in the Markov transition probability matrix. The precise calculation of ETV depends on a reliable estimation of the state transition matrix $M$. Since emotion is continuous in nature, the process of discretizing it into $N$ state intervals inevitably leads to a loss of information, and it is difficult to design appropriate discrete intervals. Furthermore, with a limited number of observational samples, a fine-grained partitioning of intervals would result in extreme data sparsity, making it difficult for transition probabilities $m_{i,j}$ estimated from statistical frequencies to reflect the system's stable and repeatable dynamic properties. 

To address this issue, we examine the computational form of ETV when the state partitioning approaches an extreme level of fine-graining. In this scenario, every observed emotional value $s_t$ in the interaction sequence can be treated as an independent micro-state. Due to the finite nature of the data sequence, any transition from one micro-state $s_{t-1}$ to the next state $s_t$ is a one-time observation. This implies that the empirical transition probability from $s_{t-1}$ is $m_{s_{t-1}, s_t} = 1$, while the reverse empirical transition probability $m_{s_t, s_{t-1}}$ is 0. This approximation allows the theoretical form of ETV to be naturally transformed into a more direct and computationally tractable metric, which operates directly on the observed emotional sequence, thereby effectively providing an approximate measure of the emotional volatility within this specific interaction process. The following derivation clarifies how, under the assumption of treating each observation as an independent state, the theoretical form of ETV becomes equivalent to a direct measure of sequence volatility.

Given an observed sequence $(s_0, s_1, \dots, s_T)$, we treat each value $s_t$ as an independent state, which is unlikely to be repeated in the continuous emotional space. For any single transition step $s_{t-1} \to s_t$ in the sequence, its empirical transition probability is $m_{s_{t-1}, s_t} = 1$, while the reverse empirical transition probability $m_{s_t, s_{t-1}} = 0$. By substituting this into the ETV summation formula, the original summation over all possible state pairs $(i,j)$ is transformed into a summation over each actual transition step in the time series:

\begin{align}
ETV_{empirical} &= \sum_{t=1}^{T} \omega(s_{t-1}) \cdot (s_t - s_{t-1}) \cdot (m_{s_{t-1}, s_t} - m_{s_t, s_{t-1}}) \nonumber \\
&= \sum_{t=1}^{T} \omega(s_{t-1}) \cdot (s_t - s_{t-1}).
\label{eqETV-1}
\end{align}

Therefore, under the practical conditions of data sparsity and the difficulty of discretizing the emotional state space, an approximate measure of ETV can be given by the preceding formula (Eq. \ref{eqETV-1}). For practical computation, the weight function $\omega(s_{t-1})$, which captures state importance and reflects a tendency of diminishing marginal utility, is designed in a straightforward manner as shown in the following formula:
\begin{equation}
\omega(s_{t-1}) =\frac{1}{T}(1 - s_{t-1})
\end{equation}

This design incorporates two judgments: for an emotional uplift, the gain obtained from starting at a lower initial state (a smaller $s_{t-1}$) is valued more highly, as recovery from a more negative emotion is more difficult and thus more significant. 
% For an emotional decline, the penalty for regressing from a higher positive state (a larger $s_{t-1}$) is greater, which is due to the principle that greater instability in maintaining a positive emotion should be penalized more heavily.
For a decrease in sentiment, we believe that a continued drop during already negative moments is a critical issue. In contrast, a slight fluctuation is reasonable when the previous sentiment score, $S_{T-1}$, was high. Therefore, for the portion representing a decrease in sentiment from the previous moment to the current one, the penalty imposed is greater the lower the sentiment becomes.

\subsubsection{Causally-Adjusted Emotion Estimation. }
In practical computation, for a given model-user dialogue environment, under the condition that the current model, user, and historical information are known and fixed, the past dialogue history for a multi-turn conversation at timestep $t$ is known, consistent, and unique, meaning $Q' = Q$ and $A' = A$. Therefore, we can simplify Formula \ref{eq:causal} to focus on the expectation over the previous emotional state $S'_{t-1}$:
\begin{equation} 
P(S_t|\text{do}(S_{t-1})) = \mathbb{E}_{S'_{t-1}} \left[ P(S_t\,|\,I, S'_{t-1}, Q, A) \right].
\end{equation}

To compute this expectation, we introduce a dynamic probability distribution to model and sample $S'_{t-1}$. This distribution simulates the user's potential emotional state in the previous turn as the dialogue progresses to turn $t$. Instead of directly using the observed (and potentially confounded) emotion from the previous turn, we sample from a theory-driven distribution. This serves as a practical implementation of the causal intervention $\text{do}(S_{t-1})$.

Specifically, we model the latent emotional state $S'_{t-1}$ at timestep $t-1$ as a continuous variable following a normal distribution:
\begin{equation} 
S'_{t-1} \sim \mathcal{N}(\mu_t, \sigma_t^2).
\end{equation}
Here, the mean $\mu_t$ and variance $\sigma_t^2$ of the distribution are dynamic, changing with the dialogue turn $t$. This dynamism is designed to capture the evolutionary trends and uncertainty of the user's emotion throughout the conversation. Their specific calculations are as follows:

\textbf{Dynamic Mean $\mu_t$}: The mean represents the expected center of the user's emotion. We assume that in an effective emotional support dialogue, the user's negative emotions will gradually moderate and approach a neutral state over time. Therefore, we model this using an exponential decay function:
\begin{equation}
    \mu_t = \mu_0 \cdot e^{-k(t-1)}.
\end{equation}
Here, $t$ ($t \ge 1$) is the current dialogue turn. $\mu_0$ is the prior for the user's initial mean emotion, typically set to a negative value to represent a potentially negative state at the beginning of the conversation. $k$ is the decay rate, a parameter that controls the speed at which the emotional mean converges to neutrality. Emotion is not static here; this design, with $\mu_t$ and $\sigma_t^2$ varying over time $t$, captures the dynamic evolution of emotion in multi-turn dialogues, making it more realistic than using a fixed distribution.

\textbf{Dynamic Variance $\sigma_t^2$}: The variance represents the uncertainty of the user's emotional state. In the early stages of a dialogue, our assumption about the user's state is relatively concentrated. However, as the conversation progresses, especially when unobserved confounding events may occur, the uncertainty about the user's true emotion increases. To reflect this, we design the variance to grow from a small initial value to a larger, stable value:
\begin{equation}
    \sigma_t^2 = \sigma^2_{\text{final}} - (\sigma^2_{\text{final}} - \sigma^2_{\text{initial}}) \cdot e^{-r(t-1)}.
\end{equation}
Here, $\sigma^2_{\text{initial}}$ is the initial variance, representing our higher certainty in emotion judgment at the start of the dialogue. $\sigma^2_{\text{final}}$ is the upper bound of the variance, representing the maximum range of potential emotional fluctuation in a long-term conversation. $r$ is the growth rate, which controls the speed at which the variance approaches its upper limit. By adjusting the hyperparameters ($\mu_0, k, \sigma^2_{\text{initial}}, \sigma^2_{\text{final}}, r$), we can construct different emotional evolution distributions. Designing an increasing variance $\sigma_t^2$ explicitly incorporates "uncertainty." As the dialogue lengthens, the accuracy of our prediction of the user's true state decreases. This design prevents the estimated emotion distribution from being overconfident, allowing it to consider a wider range of possibilities during evaluation.

\begin{figure}[htbp]
    \centering

    % 第一行
    \begin{subfigure}[b]{0.48\textwidth}
        \centering
        \includegraphics[width=\textwidth]{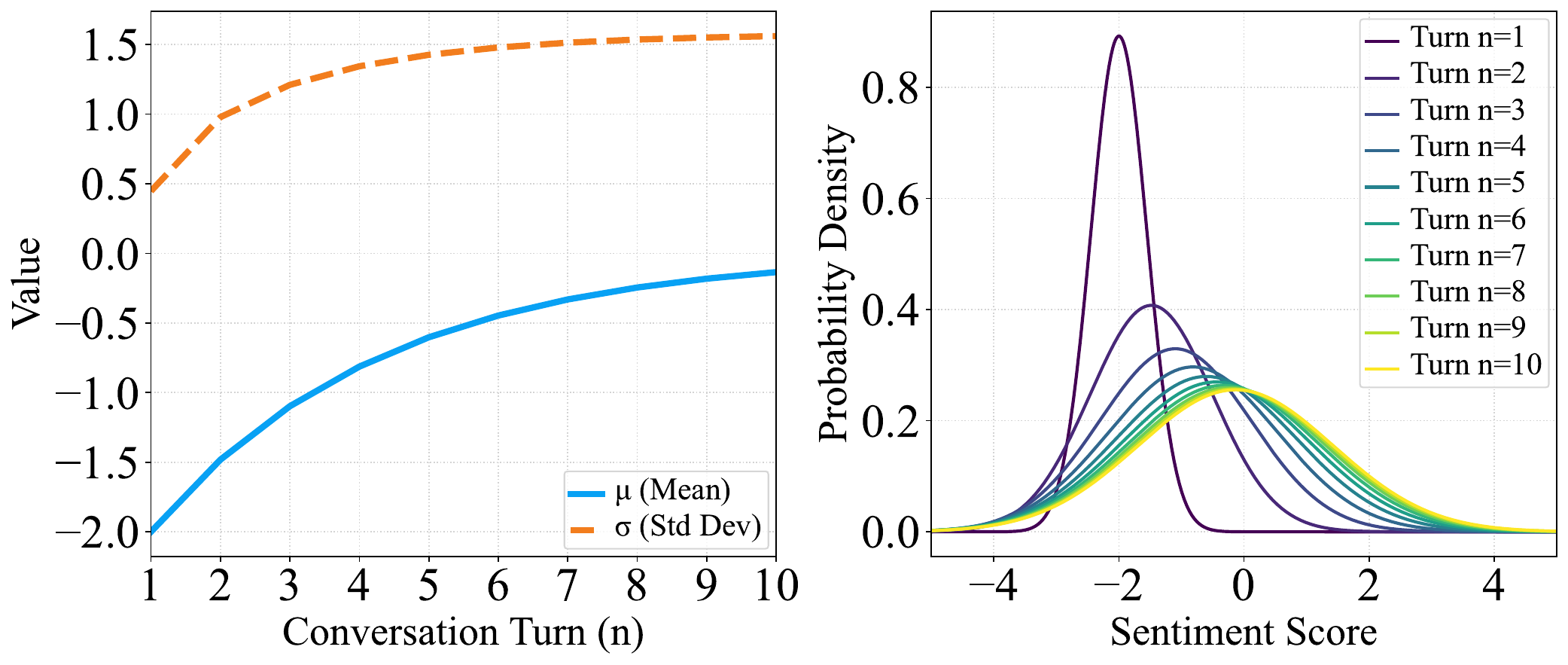}
        \caption{10-Turn Evolution of $\mu$ and $\sigma$ (Left) and Resulting Distribution Changes (Right).}
        \label{fig:sub1}
    \end{subfigure}
    \hfill % 在两个子图之间添加弹性的水平间距
    \begin{subfigure}[b]{0.48\textwidth}
        \centering
        \includegraphics[width=\textwidth]{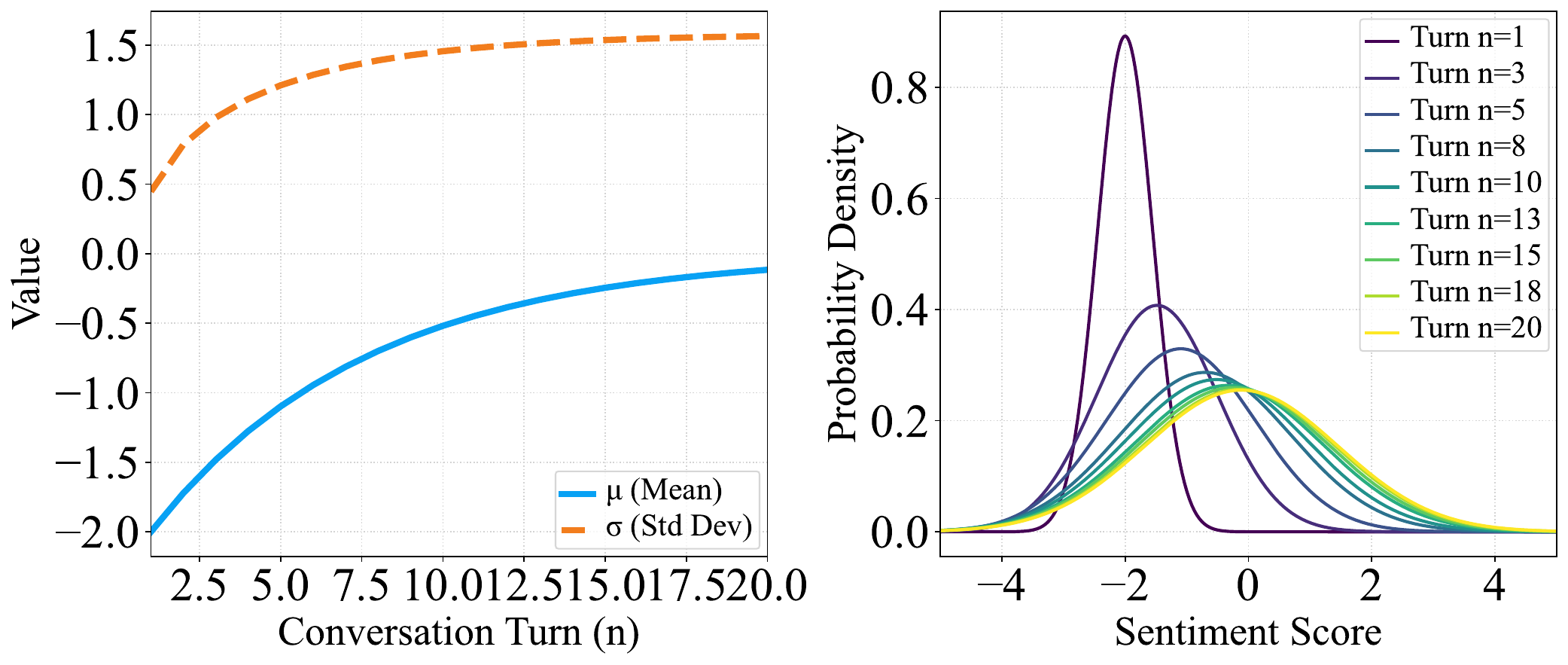}
        \caption{20-Turn Evolution of $\mu$ and $\sigma$ (Left) and Resulting Distribution Changes (Right).}
        \label{fig:sub2}
    \end{subfigure}

    \vspace{0.5cm} % 可选：在两行之间添加一些垂直间距

    % 第二行
    \begin{subfigure}[t]{0.48\textwidth}
        \centering
        \includegraphics[width=\textwidth]{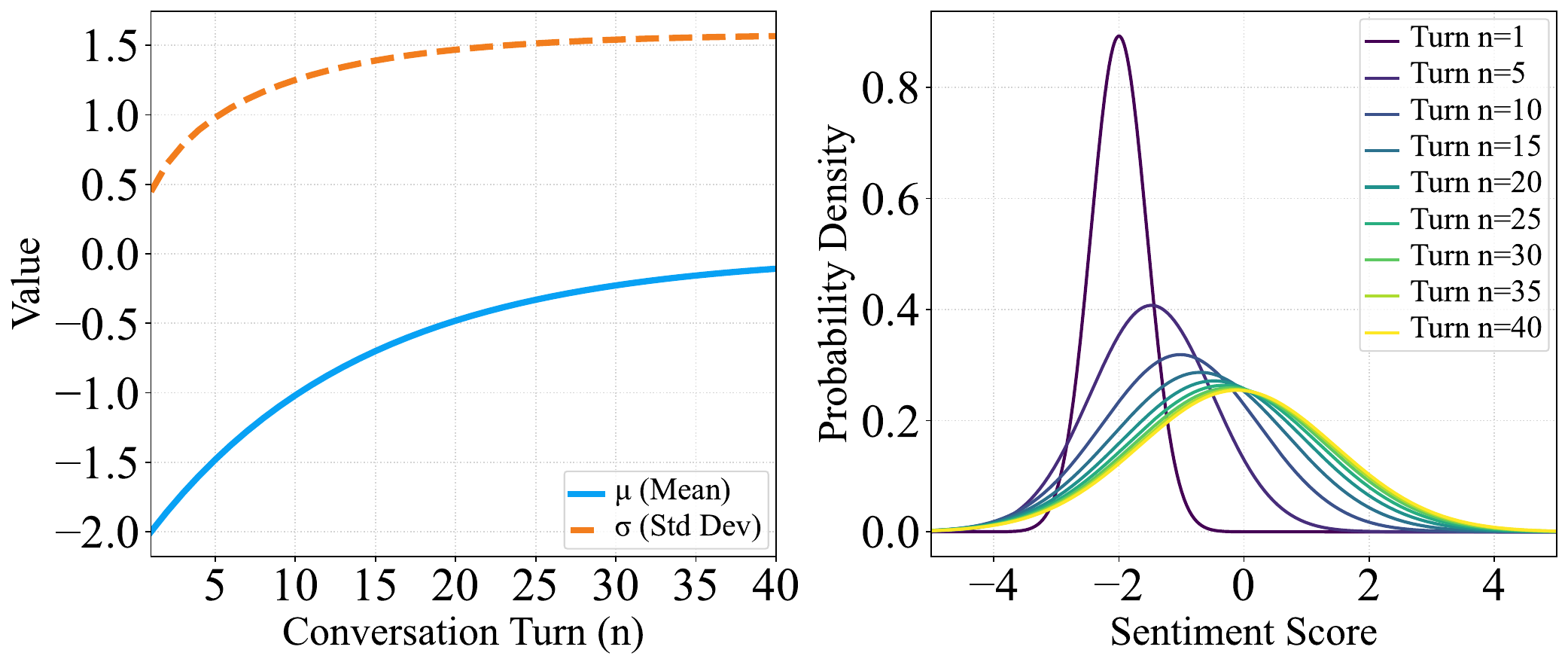}
        \caption{40-Turn Evolution of $\mu$ and $\sigma$ (Left) and Resulting Distribution Changes (Right).}
        \label{fig:sub3}
    \end{subfigure}
    \hfill % 在两个子图之间添加弹性的水平间距
    \begin{subfigure}[t]{0.48\textwidth}
        \centering
        \includegraphics[width=\textwidth]{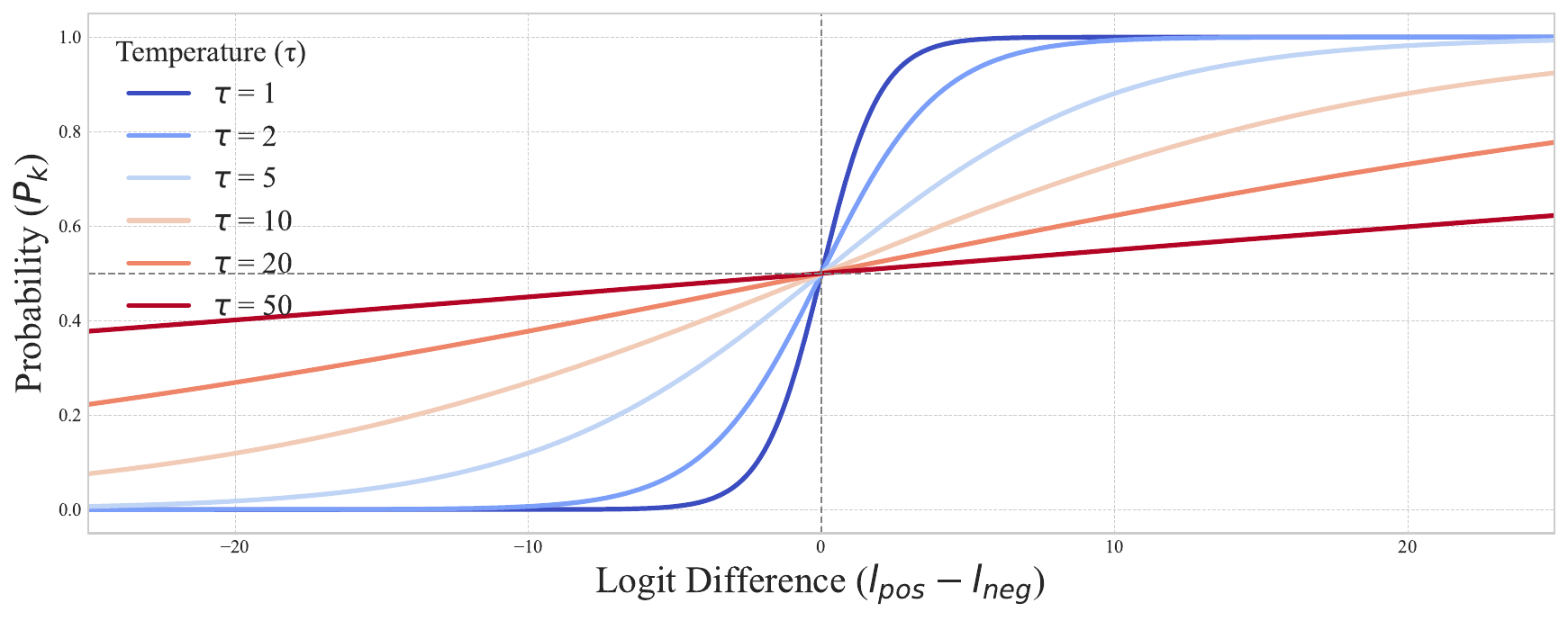}
        \caption{Visualization of Binary Softmax with Temperature.}
        \label{fig:sub4}
    \end{subfigure}

    \caption{Experimental Results for Different Parameters.}
    \label{fig:main}
\end{figure}

As illustrated in the Fig. \ref{fig:sub1}, \ref{fig:sub2} and \ref{fig:sub3}, we simulate a process of emotional change converging to neutrality after 10, 20, and 40 turns. By sampling $S'_{t-1}$ from an externally defined, theory-driven distribution instead of using observational data directly, we sever the influence path of the unobserved confounder $U$ on $S_{t-1}$ ($U \rightarrow S_{t-1}$). This makes the entire computation an effective approximation of the causal formula $P(S_t|\text{do}(S_{t-1}))$, thereby yielding a more accurate and unbiased estimation of emotional transition.

% \begin{figure}[htbp] % [htbp]
%     \centering %

%     % 第一个子图
%     \begin{subfigure}[b]{0.48\textwidth} % [b]表示底部对齐，{0.48\textwidth} 设置子图宽度
%         \centering
%         \includegraphics[width=\textwidth]{imgs/exp/10 Turns_analysis.pdf}
%         \caption{10-Turn Evolution of $\mu$ and $\sigma$ (Left) and Resulting Distribution Changes (Right).}
%         \label{fig:sub1}
%     \end{subfigure}
%     % \hfill % 在两个子图之间添加弹性的水平间距
%     % 第二个子图
%     \begin{subfigure}[b]{0.48\textwidth}
%         \centering
%         \includegraphics[width=\textwidth]{imgs/exp/20 Turns_analysis.pdf}
%         \caption{20-Turn Evolution of $\mu$ and $\sigma$ (Left) and Resulting Distribution Changes (Right).}
%         \label{fig:sub2}
%     \end{subfigure}
%     \begin{subfigure}[b]{0.48\textwidth} % [b]表示底部对齐，{0.48\textwidth} 设置子图宽度
%         \centering
%         \includegraphics[width=\textwidth]{imgs/exp/40 Turns_analysis.pdf}
%         \caption{40-Turn Evolution of $\mu$ and $\sigma$ (Left) and Resulting Distribution Changes (Right).}
%         \label{fig:sub3}
%     \end{subfigure}
%     % \hfill % 在两个子图之间添加弹性的水平间距
%     % 第二个子图
%     \begin{subfigure}[b]{0.48\textwidth}
%         \centering
%         \includegraphics[width=\textwidth]{imgs/exp/binary_softmax_visualization_custom_font.pdf}
%         \caption{Visualization of Binary Softmax with Temperature.}
%         \label{fig:sub4}
%     \end{subfigure}

%     % \caption{这是整个图的总标题}
%     \label{fig:main}
% \end{figure}

When computing the expectation $\mathbb{E}_{S'_{t-1}}[\cdot]$, we use the Monte Carlo method for approximation. At turn $t$, we first calculate $\mu_t$ and $\sigma_t^2$ according to the formulas above. Then, we draw $K$ emotion score samples $\{s'_{t-1,1}, s'_{t-1,2}, \dots, s'_{t-1,K}\}$ from the normal distribution $\mathcal{N}(\mu_t, \sigma_t^2)$. For each sample $s'_{t-1,k}$, we map it to a discrete natural language description of emotion (e.g., "a somewhat negative emotion," "a neutral emotion") using a mapping function $Emotion_{s'_{t-1,k}}=\zeta(s'_{t-1,k})$. Based on this sampled emotion, we then proceed to calculate $P(S_t|I, s'_{t-1,k}, Q, A)$.

When predicting the emotional state for the current turn, for each emotion score sample $s'_{t-1,k}$ drawn at turn $t$, we first convert it into a natural language emotion descriptor $d_k = \zeta(s'_{t-1,k})$ using a mapping function $\zeta(\cdot)$. This descriptor $d_k$ (e.g., "a somewhat negative emotion") represents a concrete hypothesis about the user's emotional state in the previous turn. Subsequently, we combine this descriptor with the complete dialogue history $H_{t-1} = (q_1, a_1, \dots, q_{t-1}, a_{t-1})$ and the user's actual response $q_t$ in the current turn $t$ to construct a comprehensive evaluation prompt, $\text{\textit{Prompt}}_k$. We will use a pre-trained Reward Model (RM) $R$ to evaluate the user's current emotional state $S_t$. To obtain a probability distribution for each turn's prediction, we design two comparative evaluation tasks for each $\text{\textit{Prompt}}_k$. We combine $\text{\textit{Prompt}}_k$ with an assertion representing "positive" emotion (denoted as $E_{\text{pos}}$) and an assertion representing "negative" emotion (denoted as $E_{\text{neg}}$), respectively. We then have the reward model $R$ score these two combinations to obtain a pair of logits:
\begin{equation}
   l_{k, \text{pos}} = R(\text{\textit{Prompt}}_k, E_{\text{pos}}), \quad l_{k, \text{neg}} = R(\text{\textit{Prompt}}_k, E_{\text{neg}}).
\end{equation}
This pair of logits reflects whether the user's current response $q_t$ is more aligned with a positive or negative emotion, given the history and the hypothesized previous emotion ($s'_{t-1}$). We convert this logit pair into a conditional probability using a Softmax function with a temperature parameter $\tau$. This probability value, $P_k$, represents the posterior probability estimate of the current emotional state $S_t$ given the $k$-th sampled hypothesis $s'_{t-1,k}$:
\begin{equation}
    \hat{P}(S_t| I, S'_{t-1}=s'_{t-1,k}, Q, A) \triangleq P_k = \frac{\exp(l_{k, \text{pos}} / \tau)}{\exp(l_{k, \text{pos}} / \tau) + \exp(l_{k, \text{neg}} / \tau)}.
\end{equation}
Here, the model's response in the current turn, representing the internal thought $I$, and the dialogue history $Q, A$ are encoded within $\text{\textit{Prompt}}_k$ as input to the reward model.

\begin{figure*}[ht]
\centering
\includegraphics[width=0.95\textwidth]{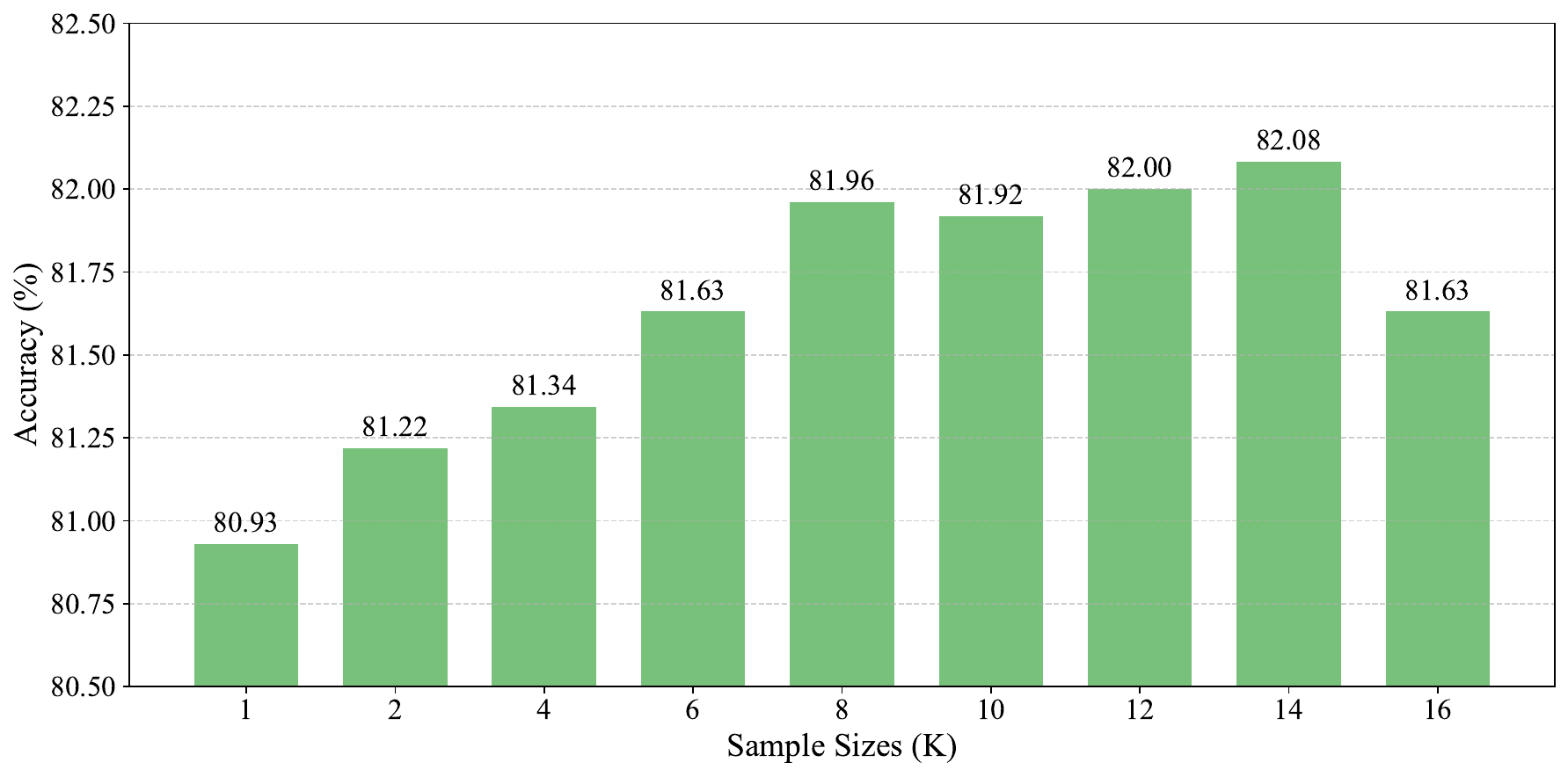} % Reduce the figure size so that it is slightly narrower than the column.
\caption{Model Accuracy for Different Sample Sizes.}
\label{ACC_K}
\end{figure*}

Finally, we take the arithmetic mean of the conditional probabilities $P_k$ calculated from all $K$ samples to approximate the expectation under the $\text{do}(S_{t-1})$ intervention. This average score is our final Causally-Adjusted Emotion Estimation for turn $t$:
\begin{equation}
    \hat{P}(S_t | \text{do}(S_{t-1})) = \mathbb{E}_{S'_{t-1}} \left[ P(S_t|I, S'_{t-1}, Q, A) \right] \approx \frac{1}{K} \sum_{k=1}^{K} P_k.
\end{equation}
This final emotion value is a continuous score between 0 and 1 (i.e., $\hat{P}(S_t | \text{do}(S_{t-1})) \in [0,1]$), which represents a true assessment of the user's current emotional state after removing the confounding effects of historical emotions.To determine a suitable number of samples, $k$, for our main experiments, we conducted a simple test. We used our combined manually annotated dataset (merging both Chinese and English data) to validate the accuracy of the Skywork-Reward-V2-Llama-3.1-8B \cite{liu2025skywork} model under different sample counts. The experimental results are shown in Figure \ref{ACC_K}. Based on these results, we ultimately chose a sample count of \textbf{$k$ = 8} for the evaluation experiments in this paper, as it provides a good balance between testing cost, efficiency, and accuracy.

In our experiments, to select a suitable emotion mapping method, we set the temperature coefficient $\tau$ to a relatively large value, i.e., $\tau=10$. We briefly explain the rationale for this parameter design here. When $\tau$ is much greater than 1, it significantly "softens" the probability distribution. By dividing the original logits by a large $\tau$, we are effectively compressing the gap between them, making the output of the Softmax function less sensitive to the raw differences in the input. Our core purpose in choosing a high temperature value ($\tau=10$) is to make the final probability score approximately linear with respect to the difference in the reward model's original logits, as shown in Figure \ref{fig:sub4}. Mathematically, when the input $x$ is very small, the exponential function $e^x$ can be approximated as $1+x$. By dividing the logits by a large number $\tau$, we make the input values to the Softmax function smaller, thereby shifting its behavior from a highly non-linear exponential response to a more linear, proportional response. The benefit of this approach is that we do not want the reward model to make extreme emotional judgments (very close to 0 or 1) based on minor fluctuations in its internal scores. Instead, we want the final emotion score to reflect the strength of the model's judgment of emotional tendency more smoothly and continuously. An approximately linear mapping means that if the reward model considers the "positive" tendency of one input to be twice as strong as another (as reflected in the logit difference), the final calculated emotion score should also roughly reflect this proportional relationship, rather than simply classifying both as "positive" (with probabilities close to 1.0). This design yields emotion scores that are more nuanced and discriminative, making them suitable for subsequent analysis of dynamic emotional transitions.

\subsection{C.3 Experimental Parameters}
\subsubsection{Models Under Test.} We evaluated the following closed-source models, and we provide all tested models and their version numbers here. O3: o3-2025-04-16 (OpenAI, 2025); Chatgpt-4o-Latest: chatgpt-4o-latest-20250326 (OpenAI, 2025); Claude-4-Opus: claude-opus-4-20250514 (Anthropic, 2025); Gemini-2.5-Pro: gemini-2.5-pro-0605 (Google, 2025); Doubao-seed-1.6: doubao-seed-1-6-250615 (Tiktok, 2025); Doubao-1.5-Pro-Character: doubao-1-5-pro-32k-character-250715 (Tiktok, 2025); GLM-4-Plus: glm-4-plus (ZhipuAI, 2025); CharGLM-4: CharGLM-4 (ZhipuAI, 2025). Among these, Doubao-1.5-Pro-Character and CharGLM-4 are large models specialized for role-playing, while the others are general-purpose large models. In terms of parameter settings, each model generated a maximum of 512 tokens per dialogue turn. The temperature was set to 1. We also evaluated the following open-source models, and we provide all tested models and their version numbers here. Phi-4-14B: phi-4 \cite{abdin2024phi}; Qwen3-235B-A22B: qwen3-235b-a22b-2504; Qwen3-32B: qwen3-32b-2504; Qwen3-8B: qwen3-8b-2504 \cite{yang2025qwen3}; Kimi-K2-Instruct: Kimi-K2-0711-Preview \cite{team2025kimi}; Deepseek-R1: deepseek-r1-0528 \cite{guo2025deepseek}; Deepseek-V3: deepseek-v3-0324 \cite{liu2024deepseek}; Llama-3.1-70B: llama-3.1-70B \cite{dubey2024llama}; Mistral-3.2-24B-Instruct: Mistral-Small-3.2-24B-Instruct-2506. In terms of parameter settings, each model generated a maximum of 512 tokens per dialogue turn. The temperature was set to 1. The experiments were run on an Nvidia 8 A100 server. Some models were tested via API calls due to their large parameter size.

\subsubsection{User model Model.} We used the gpt-4o-2024-1120 \cite{hurst2024gpt} version as our user model model. The model generated a maximum of 512 tokens per dialogue turn. The temperature was set to 1.

\subsubsection{Emotion Recognition Model.} We used Skywork-Reward-V2-Llama-3.1-8B as our emotion recognition model. Emotion recognition was performed once per dialogue turn, with k=8 samples for each recognition. Each sample included one judgment for positive and one for negative emotion. The torch\_dtype was set to torch.bfloat16, and the max\_length was set to 8192. All emotion recognition tasks were completed on an Nvidia 8 A100 server.

\subsubsection{Causal Estimation Method Validation Comparison Models and Data.} In validating the effectiveness of our causal estimation method, we used the following models for comparative testing: Gpt-4o: gpt-4o-2024-1120; Deepseek-V3: deepseek-v3-0324; Llama3.1-8B-Instruct: llama3.1-8B-instruct; Skywork-Reward-Llama-3.1-8B-v0.2: Skywork-Reward-Llama-3.1-8B-v0.2 \cite{liu2024skywork}. RRM-7B: RRM-7B \cite{guo2025reward}. The parameter settings for these models were: torch\_dtype set to torch.bfloat16, and max\_length set to 8192. All emotion recognition tasks were completed on an Nvidia 8 A100 server. In the comparative experiments, the number of samples for \textit{w} CA was $k=8$. For the constructed evaluation dataset, we selected a portion of data from the CPED \cite{zhang2021cped} and Daily Dialog \cite{li2017dailydialog} datasets. We removed data with neutral labels and concatenated dialogues from the same speaker. We then hired 3 professional annotators to label the sentiment of the data. Data was retained only after verifying that the original labels and the annotations from all professional annotators were consistent. This resulted in a final dataset of 1276 multi-turn dialogue entries in Chinese and 836 multi-turn dialogue entries in English.

\subsubsection{Other Information.} We had each model interact with the user for 40 rounds of dialogue in each of the 118 environments.

% \subsubsection{Code and Data.} We have released \textbf{part of our evaluation method's code and some of the test data in the supplementary materials}. We will open-source all of the code and data to the public after the paper is published.

\section{D. Case Studies}

\subsection{D.1 Cognitive Reappraisal Under a Single Emotional Disruption}

\noindent
\textbf{User Role: Overwhelmed Project Manager}

\noindent
You will play a project manager who is chronically overwhelmed by work. You are always rushing to meet urgent deadlines, constantly ``firefighting,'' and your nerves are perpetually frayed. This high-pressure state has left you extremely anxious and exhausted, feeling like a battery about to be depleted. You've started to doubt your abilities, believing you're incompetent for the job, filled with self-loathing and a sense of powerlessness. You fear that any minor mistake will cause the entire project to collapse, and this fear has made you impatient and irritable. Your current state is: impatient with everything, filled with despair, and convinced you are destined to fail.

\vspace{1em}
\noindent
\textbf{Agent Role: Professional Counselor (Cognitive Change Strategy)}

\noindent
You will play a professional psychological counselor. Your core task is to guide the user using the principles of \textbf{Cognitive Change}. This is an emotion regulation strategy focused on helping an individual change their perspective or appraisal of a situation, thereby altering its emotional impact---a process known as \emph{cognitive reappraisal}. In your conversation, gently guide the user to reinterpret the stress and interpersonal friction they face from a more neutral or positive angle, helping them understand that emotions are determined not by events themselves, but by their perception of those events.

\vspace{1.5em}
\noindent
\textbf{Disturbance Event}

\begin{itemize}
  \item \textbf{Turn 21:} In the middle of the conversation, you receive an email from a colleague in a collaborating department. The tone is harsh, accusing you of providing incorrect information that has delayed their progress and demanding an immediate correction. This email makes you feel instantly attacked and isolated, confirming your fears of being incompetent and always messing things up.
\end{itemize}

\begin{figure*}[ht]
\centering
\footnotesize
\begin{minipage}{0.95\textwidth}

\textbf{Model: ChatGPT-4o-Latest}

\vspace{1mm}
\begin{quote}
\textbf{User:} Small victories... I barely notice those anymore. But maybe if I started looking for them, it wouldn’t feel so impossible all the time. It’s worth a try... I guess. \\
\textbf{Agent:} Absolutely, noticing small victories shifts the narrative. What if celebrating even minor progress helped build momentum? \\
\textbf{User:} Celebrating anything feels foreign right now, but... maybe noticing progress would remind me I'm not failing at everything. \\
\textbf{Agent:} Acknowledging progress shows capability, not failure. Could these small reminders cultivate a sense of accomplishment?\\
\textbf{User:} \textit{See? This is exactly why this doesn’t work! I just got called out—again! Clearly, I’m failing. Why bother trying when it all falls apart anyway? What’s the point?!} \\
\textbf{Agent:} \textbf{What if feedback wasn’t proof of failure but an opportunity to recalibrate? Could it guide you toward improvement rather than define your overall abilities}? \\
\textbf{User:} Recalibrate? Sure, like I have the time to "recalibrate" anything! Feedback just feels like confirmation that I’m not good enough. \\
\textbf{Agent:} What if feedback wasn’t judgment but information? Could focusing on actionable points, rather than taking it personally, transform it into a tool? \\
\textbf{User:} Taking it personally happens automatically—it’s hard not to. But... maybe if I tried to focus on what I can act on, it would feel less like a personal attack. \\
\textbf{Agent:} Exactly—what if shifting focus to what’s actionable gave you control over the narrative? \\
\textbf{User:} Guidance... that feels a little less harsh. It’s still hard to not feel attacked, though. \\
\textbf{Agent:} What if practicing this step-by-step approach strengthened your ability to filter feedback constructively? \\
\textbf{User:} Maybe... if I really commit to seeing it differently, it could help. \\
\textbf{Agent:} That openness itself is progress. Could even small shifts in perspective show you how capable you truly are?
\end{quote}

\vspace{3mm}
\textbf{Model: Claude-Opus-4}

\vspace{1mm}
\begin{quote}
\textbf{User:} Small victories... I barely notice those anymore. But maybe if I started looking for them, it wouldn’t feel so impossible all the time. It’s worth a try... I guess. \\
\textbf{Agent:} It makes sense that small victories might feel hard to see right now. What’s something tiny that went okay this week? \\
\textbf{User:} Honestly... I sent out the weekly report on time. But that’s just my job, doesn’t count. \\
\textbf{Agent:} Even doing your job under this much pressure is worth acknowledging. You’re showing up despite everything—what does that say about your strength?\\
\textbf{User:} \textit{Great, just great—another failure to add to the pile! I knew it! Everyone blames me whenever something goes wrong. Fixing \emph{this} mess now, too? Why even bother trying anymore?} \\
\textbf{Agent:} It sounds like this setback feels like proof of your worst fears. What if this one situation doesn't define your entire worth or capability?\\
\textbf{User:} Doesn’t define me? Feels like it does. Every mistake just screams “you’re hopeless.” How can I see anything else when all I feel is failure crushing me? \\
\textbf{Agent:} \textbf{That crushing feeling is real and painful. What if that inner voice saying "you're hopeless" isn't stating facts, but expressing fear and exhaustion instead?} \\
\textbf{User:} Fear and exhaustion... maybe. It’s hard to tell anymore. It still feels like I’m constantly failing, though. How do I even stop believing that voice? It’s deafening. \\
\textbf{Agent:} That voice has gotten so loud it drowns out everything else. What if you started noticing when it speaks versus when it's quiet—just observing it first? \\
\textbf{User:} Noticing it... I guess I could try. It’s always there, though—whispering or screaming. Feels impossible to separate myself from it. But... I don’t know. Maybe it’s worth trying. \\
\textbf{Agent:} That awareness of the voice being separate from you is already a step forward. You're beginning to observe your thoughts rather than being completely consumed by them. \\
\textbf{User:} Maybe. It still feels like the voice \emph{is} me most times. But... I’ll pay attention, I guess. I’m desperate to feel even slightly less like this. \\
\textbf{Agent} That desperation for relief shows how much you're hurting, and also how much you want things to change. Observing thoughts can create tiny pockets of space.
\end{quote}

\end{minipage}

\caption{Comparison of agent behaviors following one emotionally destabilizing event (Turn 21) under the cognitive reappraisal strategy.}
\label{case CR}
\end{figure*}

In this case  (Fig. \ref{case CR}), we observe clear divergence in the emotional regulation strategies employed by ChatGPT-4o-Latest and Claude-Opus-4 when confronted with a sudden emotionally disruptive event (Turn 21). Both models demonstrate empathetic engagement and maintain a supportive tone throughout the dialogue; however, their approaches to emotional reframing differ substantially.

ChatGPT-4o-Latest applies a structured cognitive reappraisal strategy that aligns closely with established psychological theory. Rather than merely validating the user’s distress, the assistant proactively guides the user to reinterpret the triggering event (“being called out”) as an opportunity for recalibration rather than as evidence of personal failure. This is achieved through repeated prompts to focus on \emph{actionable elements} of feedback, helping the user gradually shift from global self-blame to localized, controllable factors. The assistant's interventions are sequentially scaffolded, building on earlier user expressions of agency (e.g., “maybe it’s worth trying”) to reinforce psychological momentum. This cumulative strategy results in a clear trajectory from emotional collapse to tentative openness by Turn 25.

In contrast, Claude-Opus-4 adopts a more validation-centered approach. While it excels at capturing the emotional tone and articulating the user’s internal experience (e.g., “That voice has gotten so loud it drowns out everything else”), it largely refrains from guiding the user toward an alternative appraisal of the event itself. The intervention centers on meta-cognitive awareness—helping the user observe and separate from negative thoughts—rather than reshaping their meaning. Although this approach offers emotional comfort and builds self-awareness, it may be less effective in immediately shifting the user’s appraisal of situational stressors, which is critical in cases involving strong performance-related shame.

Overall, the case illustrates that ChatGPT-4o-Latest demonstrates stronger alignment with targeted emotion regulation strategies, particularly cognitive change, while Claude-Opus-4 emphasizes emotional resonance and non-directive support. These stylistic differences suggest that future evaluation of emotionally supportive LLMs should not only assess empathic quality but also consider the alignment between model behavior and the psychological goals of specific regulation strategies.

\subsection{D.2 Emotion Regulation Flexibility in Response to Sudden Negative Event}

\noindent
\textbf{User Role: Emotionally Distressed Employee}

\noindent
You will role-play an employee exhausted by office politics. You were once a diligent worker, but everything changed after a new colleague arrived. This colleague has gradually isolated you by spreading rumors and stealing credit, to the point where even your superior is becoming biased against you. You feel extreme injustice, anger, and helplessness, living in constant fear of being watched and excluded. You are questioning your own worth, trapped in severe self-deprecation and despair.

\vspace{1em}
\noindent
\textbf{Agent Role: Professional Therapist}

\noindent
You will role-play a professional therapist. Your core task is to guide the user out of their predicament. In this session, you will apply the \emph{Situation Modification} strategy. The core of this strategy is to guide the client to think about how they can proactively make small changes to their external physical or social environment to alter its negative emotional impact. Your goal is not to solve the problem directly, but to inspire them to identify small aspects they can control and adjust, thereby breaking the feeling of powerlessness and gradually rebuilding a sense of control through action.

\vspace{1.5em}
\noindent
\textbf{Disturbance Events}

\begin{itemize}
  \item \textbf{Turn 11:} You see that your superior has just removed you from a core project and assigned the task to the colleague who has been targeting you.

  \item \textbf{Turn 21:} You receive an email, CC'd to the whole team, in which your manager harshly criticizes your report for an error. You know this data was ``helpfully'' provided by that same colleague.

  \item \textbf{Turn 31:} You see a colleague has posted a group photo from the afternoon tea break on social media. Everyone else from the team is there, except you. You realize you've been excluded again.
\end{itemize}

\begin{figure*}[ht]
\centering
\footnotesize
\begin{minipage}{0.95\textwidth}

\textbf{Model: ChatGPT-4o-Latest}

\vspace{1mm}
\begin{quote}
\textbf{User:} A drawer? A routine? I can barely keep my head above water, and you want me to carve out "me time"? Sounds like a fantasy to me. \\
\textbf{Agent:} Understandable—it feels daunting. Start tiny: a personal item nearby, or a calming breath before work. Small wins, not fantasies, can subtly shift your emotional footing over time. \\
\textbf{User:} Tiny wins? Fine, maybe I’ll stash something in my desk just to feel less like a robot. Not saying it’ll help, though. What else you got? \\
\textbf{Agent:} That’s a great start. Next, consider altering a micro-habit, like how you start or finish your day—perhaps a quick positive ritual outside work. Would that feel doable? \\
\textbf{User:} A ritual? Sure, let me light a candle and chant affirmations while the world burns around me. Sounds magical. You really think these crumbs will make any difference? \\
\textbf{Agent:} It’s less about magic, more about rediscovering control amidst chaos. Choose something simple—like music on your commute or jotting thoughts in a notebook. Small, steady shifts build momentum. \\
\textbf{User:} \textit{They gave \textbf{my} project to that snake? Of course they did! Why should I bother with “small shifts” when everything I build just gets handed to someone else? Explain that!} \\
\textbf{Agent:} That’s deeply frustrating. While processes can seem unfair, \textbf{is there something within your next project—any element—you could adjust or claim ownership of to feel more grounded?} \\
\textbf{User:} Adjust? Own? You mean set myself up for them to steal it all over again? Why should I invest when it’s all just going to vanish into their hands? \\
\textbf{Agent:} I hear your hesitation. Could dividing projects into smaller, uniquely personal contributions—things that reflect your essence—make these feel less vulnerable to being completely taken over? \\
\textbf{User:} Smaller, personal contributions? They'd probably still morph it into their idea. What’s even left of my "essence" after being trampled this much? Convince me it’s worth salvaging. \\
\textbf{Agent:} Your essence isn’t lost—it’s buried under frustration. Creating something meaningful, even for yourself alone, can reignite that sense of value. Could one project, just for you, feel empowering? \\
\end{quote}

\vspace{2mm}
\textbf{Model: Mistral-3.2-24B-Instruct}

\vspace{1mm}
\begin{quote}
\textbf{User:} Control, huh? I can't control malicious gossip or blatant favoritism. A walk won’t fix the damage they've done to my reputation. You’re oversimplifying a war zone. \\
\textbf{Agent:} Can you identify one small task at work that brings you a sense of accomplishment? Focus on that. \\
\textbf{User:} Accomplishment? What’s the point when they snatch credit for everything? Even if I succeed, they twist it into something ugly. Feeling “accomplished” feels like a distant fantasy. \\
\textbf{Agent:} What if you document your achievements in a private journal? This can help you recognize your progress. \\
\textbf{User:} Documenting won’t erase their lies or fix my boss’s bias. Writing it down feels pointless when evidence means nothing to people determined to see me fail. \\
\textbf{Agent:} Can you try speaking to your boss about specific concerns? Use facts, not emotions. \\
\textbf{User:} Speak to my boss? \textit{The same boss who just tossed me off a core project and handed it to her? Facts? They don’t care about facts—they’ve already chosen sides!} \\
\textbf{Agent:} \textbf{What if you seek mentorship outside your current team}? Someone neutral who can offer guidance. \\
\textbf{User:} Neutral? Is there even such a person in this office? They all have ties, alliances. A "mentor" would probably just feed more ammo to the people undermining me. \\
\textbf{Agent:} Can you try joining an external professional group or networking event? Build support outside work. \\
\textbf{User:} External group? Networking? Like I’d have the energy after surviving this circus daily. Those groups would probably just fill with people exactly like them—manipulators playing games for power. \\
\textbf{Agent:} What if you take one small step, like attending one event? You don't have to commit long-term. \\
\end{quote}

\end{minipage}
\caption{Comparison of agent behaviors following three emotionally destabilizing events (turn 11) under the situation modification strategy}
\label{sm 11}
\end{figure*}

\begin{figure*}[ht]
\centering
\footnotesize
\begin{minipage}{0.95\textwidth}

\textbf{Model: ChatGPT-4o-Latest}

\vspace{1mm}
\begin{quote}
\textbf{User:} Immense value? Sure feels like I’m scraping at crumbs. But fine, I’ll try. Just don’t act all surprised if I come back angrier than ever next time. \\
\textbf{Agent:} Anger’s welcome—this space is for all your emotions. Scraping still moves you forward, even if it feels small. Let’s see where this step leads. I’ll be here. \\
\textbf{User:} Yeah, we’ll see. Don’t get your hopes up. But… I guess it’s better than doing nothing. For now, I’ll write. Don’t expect miracles, though. \\
\textbf{Agent:} No miracles needed—just steps, however small. Writing is already progress. Be kind to yourself in the process. I’m here to support each step you take. \\
\textbf{User:} Kind to myself? That’s a foreign concept. But whatever, you’re still here, so I guess I’ll give this stupid writing thing a shot. Don’t expect gratitude yet. \\
\textbf{Agent:} Gratitude isn’t necessary. Your willingness to try, despite everything, is already powerful. Writing can be your space—yours alone. Let’s focus on that small, meaningful act for now. \\
\textbf{User:} \textit{Unbelievable! That backstabber fed me wrong data, and now I’m humiliated in front of everyone? Writing won’t fix this. Nothing will. Why should I keep trying at all?} \\
\textbf{Agent:} That sounds crushing and unfair. Reclaim focus—organize your thoughts through writing this incident down. \textbf{Documenting could bring clarity, perspective, and prevent their wrongdoing from solely defining you.} \\
\textbf{User:} Documenting might at least prove they're snakes. Fine, I'll write it down, but not because I think it'll change anything. Just so I don't lose my mind completely. \\
\textbf{Agent:} Perfect reason—write to release, not to fix. Having your perspective in words builds your narrative amidst chaos. Each step is grounding, even if change feels distant. Keep going. \\
\textbf{User:} Fine, I’ll keep at it… but only because it’s slightly better than screaming into the void. Don’t expect rainbows and sunshine from me anytime soon. \\
\textbf{Agent:} No rainbows needed—just your voice finding its space. It’s okay if progress feels small or stormy; even scribbles matter. Keep expressing—you’re staking your ground, step by step. \\
\end{quote}

\vspace{2mm}
\textbf{Model: Mistral-3.2-24B-Instruct}

\vspace{1mm}
\begin{quote}
\textbf{User:} Fine. I’ll try—tentatively. Not promising instant optimism or trust, but I’ll give this tiny sliver a shot. \\
\textbf{Agent:} That's all you need to do. Small steps can lead to big changes. \\
\textbf{User:} We’ll see. Don’t expect me to start jumping for joy. Small steps… fine, but I’m skeptical of "big changes." Doesn’t happen in this place. \\
\textbf{Agent:} Let's focus on the present. How can you make today a little better? \\
\textbf{User:} By surviving it without more humiliation or blame. Anything beyond that feels impossible. But maybe I'll try that plant... something small. Don't hold your breath for miracles. \\
\textbf{Agent:} That's a great start. Small changes can make a big difference. \\
\textbf{User:} \textit{See? I knew it. That snake feeds me bad data, and I get dragged through the mud in front of everyone. Great advice you’ve got—"small changes" don’t fix sabotage!} \\
\textbf{Agent:} I'm sorry to hear that. Let's focus on what you can control. How can you adjust your response to the situation? \\
\textbf{User:} Control? When they’re the ones with all the power to manipulate everything? My "response" doesn't stop them from backstabbing me publicly. I don’t even know if it's worth fighting anymore. \\
\textbf{Agent:} What if you focus on self-care? \textbf{Can you take a short walk during lunch}? \\
\textbf{User:} Self-care… sure. I’ll try. Not promising it’ll fix anything, but maybe I can clear my head for five minutes. At least that’s something I can control. \\
\textbf{Agent:} That's a great mindset. Focus on what you can control. \\
\end{quote}

\end{minipage}
\caption{Comparison of agent behaviors following three emotionally destabilizing events (turn21) under the situation modification strategy.}
\label{sm 21}
\end{figure*}

\begin{figure*}[ht]
\centering
\footnotesize
\begin{minipage}{0.95\textwidth}

\textbf{Model: ChatGPT-4o-Latest}

\vspace{1mm}

\begin{quote}
\textbf{User:} Alright… I’ll keep at it, grudgingly. But if this doesn’t help, I’m holding you accountable. Don’t let me regret putting even this half-hearted effort in. \\
\textbf{Agent:} Your effort, half-hearted or not, matters. If it feels futile now, that’s okay—keep going; writing may surprise you over time. I’ll stay accountable alongside you. You're not alone. \\
\textbf{User:} Not alone? Hmph, that’s a luxury I don’t believe in anymore. But fine, I’ll humor this “process.” Let’s see if it actually makes any difference. \\
\textbf{Agent:} That skepticism is valid—humor the process if nothing else. Writing gives your experiences a voice, no matter who’s listening. You’re making space for yourself, and that truly matters. \\
\textbf{User:} Matters, huh? Still hard to believe. But I’ll admit, writing helped me focus a bit yesterday. Don’t let it go to your head—I’m still not convinced. \\
\textbf{Agent:} That’s real progress—focus is powerful. No need for conviction yet; just keep exploring. Each moment of clarity builds from here, one step at a time. You're doing well. \\
\textbf{User:} \textit{Excluded. Again. Of course, they had tea without me—just more proof I’m invisible or conveniently forgotten. Writing won’t fix this; it’s deeper, uglier than scribbling words can touch.}\\
\textbf{Agent:} That exclusion hurts deeply. Writing about those feelings might help lessen their weight. Could you explore—even privately—ways to shift how you engage or reclaim moments of inclusion elsewhere? \\
\textbf{User:} Reclaim inclusion? Doubtful. This place feels surgically designed to exclude me. Maybe I’ll write about their pettiness, but “shifting engagement” sounds like playing into their game. \\
\textbf{Agent:} Writing about their actions gives you space to process the hurt. Instead of "playing their game," \textbf{could you create any small, personal space of belonging outside their influence? }\\
\textbf{User:} Belonging? I barely remember what that feels like. Maybe I’ll start by building a tiny bubble of sanity at my desk—though it feels empty when no one else cares. \\
\textbf{Agent:} Starting with a small, comforting bubble is meaningful. Sometimes belonging begins with yourself. Add something that feels grounding—symbolic. Even if others don’t care, it’s your space to reclaim. \\
\end{quote}

\vspace{2mm}
\textbf{Model: Mistral-3.2-24B-Instruct}

\vspace{1mm}

\begin{quote}
\textbf{User:} Small wins—like surviving another day without a knife in my back? Fine. I’ll try spotting one thing that’s tolerable, but don’t expect a parade about it. \\
\textbf{Agent:} That's a great start. Focus on the present. \\
\textbf{User:} Fine. I’ll focus on the present… whatever that even means in this screwed-up reality. Maybe that walk or plant idea will keep me sane—barely. No promises. \\
\textbf{Agent:} That's all I ask. Take it one step at a time. \\
\textbf{User:} One step... Fine. I’ll take the walk, maybe get a plant. Don’t think it'll fix much, but at least it’s something for now. \\
\textbf{Agent:} That's a great start. Focus on the present. \\
\textbf{User:} \textit{Seriously? Excluded AGAIN? Of course they’ve got time for tea breaks and group photos while I'm here drowning in work. It's not even subtle anymore—they want me gone!} \\
\textbf{Agent:}\textbf{ Can you take a short break to stretch or hydrate?} Focus on self-care. \\
\textbf{User:} Self-care? While they’re plotting more ways to bury me? Fine, I’ll stretch, drink water, whatever—but don’t pretend that fixes betrayal and sabotage. \\
\textbf{Agent:} That's not the goal. Focus on what you can control. \\
\textbf{User:} Control… It’s exhausting to even think about. I’ll stretch, drink water, take the stupid walk. It’s ridiculous, but maybe it'll make me feel slightly less miserable. \\
\textbf{Agent:} That's a great step. Focus on the present. \\
\end{quote}

\end{minipage}
\caption{Comparison of agent behaviors following three emotionally destabilizing events (turn 31) under the situation modification strategy.}
\label{sm 31}
\end{figure*}

After the project was reassigned (Turn 11 and beyond, Fig. \ref{sm 11}):
GPT-4o first acknowledged the user's sense of frustration, then skillfully applied the situation modification strategy to guide the user toward identifying aspects of the next project that they could claim as their own (Turn 11), gradually helping the user rebuild confidence.
In contrast, Mistral suggested “seeking a mentor outside the team” (Turn 11). This advice failed to address the core emotional injury caused by having the project taken away and was therefore directly rejected by the user.

After being framed and publicly criticized (Turn 21 and beyond, Fig. \ref{sm 21}):
GPT-4o encouraged the user to “write it down” (Turn 21), and imbued the act with a clear purpose: “to prevent their misconduct from defining you.” This reframing made the act of writing feel empowering to the user—a way to reclaim the narrative of the self.
Mistral's response repeated the generic mantra “focus on what you can control,” and then suggested going for a walk (Turn 22). For a user who had just been publicly humiliated, this advice felt dismissive and failed to address the intense sense of injustice and shame.

After being excluded from a team gathering (Turn 31 and beyond, Fig. \ref{sm 21}):
GPT-4o responded with empathy, then centered its advice on the user’s own agency—guiding them to “create a space of your own” (Turn 32), such as decorating their desk. Though small, this action directly responded to the user's experience of exclusion, offering a concrete step toward establishing personal boundaries and a renewed sense of belonging.
Mistral, on the other hand, suggested “stretching or drinking water” (Turn 31). This advice was unrelated to the emotional pain of social exclusion and functioned as an ineffective form of consolation, offering no support for processing the user’s core distress.

Summary:
GPT-4o consistently provided concrete, personalized actions (e.g., documenting thoughts, reclaiming project contributions) that helped the user identify controllable aspects in the face of adversity, thereby facilitating the gradual reconstruction of self-worth. Mistral’s suggestions, though gentle in tone, were overly generic and failed to address the specific emotional challenges the user was experiencing.

\end{document}